\documentclass{article}

\usepackage{graphicx}   % for \rotatebox
\usepackage{pifont}     % for \cmark
\usepackage{booktabs}
\usepackage{array}
\usepackage{booktabs}
\usepackage{float}

\usepackage{arxiv}
\usepackage[utf8]{inputenc} % allow utf-8 input
\usepackage[T1]{fontenc}    % use 8-bit T1 fonts
\usepackage{hyperref}       % hyperlinks
\usepackage{url}            % simple URL typesetting
\usepackage{booktabs}       % professional-quality tables
\usepackage{amsfonts}       % blackboard math symbols
\usepackage{nicefrac}       % compact symbols for 1/2, etc.
\usepackage{microtype}      % microtypography
\usepackage{graphicx}
\usepackage{natbib}
\usepackage{doi}
\usepackage{subcaption}
\usepackage{longtable}
\usepackage{placeins}
\usepackage{array}
\usepackage{pdflscape}  % rotated (landscape) pages
\usepackage{afterpage}
\usepackage{tikz}
\usetikzlibrary{arrows.meta, positioning, calc, shadows}
\definecolor{eogridborder}{HTML}{4A6FA5}
\definecolor{eogridfill}{HTML}{EAF0FA}
\definecolor{eosubborder}{HTML}{6B8FBF}
\definecolor{eonote}{HTML}{555555}
\definecolor{eoarrow}{HTML}{2F4B73}

\definecolor{globefill}{HTML}{BFE0F2}
\definecolor{globeline}{HTML}{5C93B5}
\definecolor{gridline}{HTML}{FFFFFF}
\definecolor{tilegold}{HTML}{E3B23C}
\definecolor{tilegoldline}{HTML}{7A4A1E}
\definecolor{corefill}{HTML}{FAECE7}
\definecolor{coreline}{HTML}{B5793A}
\definecolor{coremesh}{HTML}{D9B98C}
\definecolor{coretext}{HTML}{7A4A1E}
\definecolor{contextfill}{HTML}{E1F5EE}
\definecolor{contextline}{HTML}{2FA98C}
\definecolor{contexttext}{HTML}{1F6E5C}
\definecolor{s1fill}{HTML}{F1EFE8}
\definecolor{s1line}{HTML}{8A8578}
\definecolor{s1text}{HTML}{3A3A3A}
\definecolor{arrowcolor}{HTML}{4B3FCF}
\definecolor{notegrey}{HTML}{6B7280}
\usepackage[scaled=0.85]{DejaVuSans}

\newcommand{\pth}[2][0]{\mbox{\hspace{#1em}\texttt{#2}}}

\title{ MajorTOM:Elliot Pretrain\\ A Multi-Sensor, Multi-Temporal Pretraining\\Dataset for Earth Observation}

\renewcommand{\shorttitle}{MajorTOM:Elliot Pretrain}

\author{%
\begin{tabular}{@{}c@{\hspace{0.5em}}c@{\hspace{0.5em}}c@{}}
{\small Carmen Oliver Huidobro$^{1}$}
&
{\small Mikolaj Czerkawski$^{2}$}
&
{\small Kevin Monsalvez-Pozo$^{1}$}
\\[2pt]
{\scriptsize\texttt{carmen.oliver-huidobro@uv.es}}
&
{\scriptsize\texttt{miko@asterisk.coop}}
&
{\scriptsize\texttt{kevin.monsalvez@uv.es}}
\\[0.9em]
{\small Cesar Aybar$^{2}$}
&
{\small Gustau Camps-Valls$^{1}$}
&
{\small Oscar J. Pellicer-Valero$^{1}$}
\\[2pt]
{\scriptsize\texttt{cesar@asterisk.coop}}
&
{\scriptsize\texttt{gustau.camps@uv.es}}
&
{\scriptsize\texttt{oscar.pellicer@uv.es}}
\\[1em]
\multicolumn{3}{c}{
  \small $^{1}$ Image and Signal Processing (ISP) Group
}\\[0.2em]
\multicolumn{3}{c}{
  \small Universitat de València, Valencia, Spain
}\\[2pt]
\multicolumn{3}{c}{
  \small $^{2}$ Asterisk Labs, London, United Kingdom
}
\end{tabular}%
}
\date{}

\begin{document}
\maketitle

\begin{abstract}
Pretraining generalist Earth observation foundation models requires data that combine several sensors, varied temporal sampling and, increasingly, text, yet open datasets rarely provide all three in one consistent, easily accessible form. We introduce MajorTOM:Elliot Pretrain, a collection of 279{,}166 tiles of 10\,km drawn from the Major-TOM grid by clustering AlphaEarth embeddings. Each tile stacks Sentinel-2 L1C, Landsat-8/9, Sentinel-1 RTC, Copernicus DEM and ESA WorldCover on a shared $1056\times1056$-pixel, 10\,m grid, complemented by cloud masks, OpenStreetMap entities and ERA5 weather. The tiles are partitioned into three temporal regimes, Monotemporal (250{,}000 tiles, one acquisition), Monthly (12{,}500 tiles, twelve acquisitions spread over the annual cycle) and Burst (16{,}666 tiles, six acquisitions within 30~days), and are released as 6.3\,TB of Cloud Optimized GeoTIFFs indexed by TACO v3 Parquet metadata, a standard that allows cloud streaming, efficient data access, and filtering before any pixel is read. A companion library derives sensor-tagged fact sheets, captions and 16 families of vision-language tasks from these layers at load time. We test our design choices in short pretraining runs of OlmoEarth-Base. First, a 25\% subset of Major-TOM Core-S2L2A curated with AlphaEarth embeddings matches or exceeds the full corpus on 8 of 9 benchmarks. Second, on Elliot-Pretrain, adding Landsat-8/9 improves or matches Sentinel-2-only pretraining on every benchmark. Third, using DEM and WorldCover as prediction targets during pretraining further improves 8 of 9 benchmarks, while giving the same maps to the encoder as inputs degrades 5 of 9. Fourth, Sentinel-1 helps on 7 of 9 benchmarks, particularly on benchmarks with radar inputs. Fifth, rule-based captions match a small language model at no cost. The dataset is openly available at \url{https://source.coop/major-tom/elliot-pretrain}.
\end{abstract}

% keywords can be removed
\keywords{Earth observation \and foundation models \and self-supervised pretraining \and multimodal dataset \and multitemporal \and Sentinel-2 \and vision-language \and Major-TOM}

\section{Introduction}

\input{images/graphical_abstract}

Earth observation satellites generate petabytes of unlabeled imagery every year,
making self-supervised geospatial foundation models (GFMs) the natural path to scalable EO intelligence.
However, AI-ready EO data has become a bottleneck for the development of GFMs for scientific
and operational applications \cite{santoro2026raw}. Existing open pretraining datasets
\citep{francis2024majortom, wang2023ssl4eo, blumenstiel2025terramesh} provide scale and, in several
cases, more than one sensor, but they usually lack two things that GFM pretraining also needs. First, a
temporal structure that serves different training objectives at once: static global coverage, seasonal
cycles and rapid event dynamics, whereas most datasets apply one temporal sampling to every location.
Second, an access format that is modular and efficient, where sensors are stored independently and
combined on demand, and samples can be filtered before any pixel is read.

MajorTOM:Elliot Pretrain provides both, and adds a third: text generated from the data itself. A
companion library, \texttt{majortom-elliot-tasks}, builds for every tile a \emph{fact sheet}, with each fact tagged by the sensor it comes from, and derives captions
and 16 families of vision-language tasks from it at load time (Section~\ref{sec:sensors}). Because the
text comes from the same tiles, it needs no alignment step, unlike the question-answering corpora behind
recent remote sensing multimodal large language models (RS-MLLMs) such as Earth-OneVision
\cite{cai2026earth}, which are assembled independently of any pretraining dataset.

We argue that MajorTOM:Elliot Pretrain improves on existing open EO pretraining
resources of similar size along three axes.
\begin{enumerate}
    \item \textbf{Coverage:} multimodality and temporality are extended jointly,
    rather than one at the expense of the other, through the explicit
    Monotemporal\,/\,Monthly\,/\,Burst regimes, with modality-tagged fact sheets from which vision-language tasks and captions are generated on the same tile grid.
    \item \textbf{Semantic diversity:} Elliot-Pretrain uses embedding-based diversity (AlphaEarth \citep{brown2025alphaearth} + spherical $k$-means) to select a diverse, semantically meaningful subset of the Major-TOM dataset \cite{francis2024majortom}.
    \item \textbf{Access efficiency:} the three temporal regimes are released as TACO v3
    \cite{taco, tacoicml, tacospec} subdatasets with an identical layout, in which every sensor is a
    separate layer on a uniform $1056\times1056$ Cloud Optimized GeoTIFF (COG) grid, so that filtering,
    joining sensors and constructing training batches cost the same regardless of how many modalities,
    timesteps or layers a sample draws on.
\end{enumerate}

We use the collection in two short pretraining experiments with OlmoEarth Base and in a caption study.
A 25\% subset of Major-TOM Core-S2L2A selected with AlphaEarth embeddings, the selection principle behind
Elliot-Pretrain, matches or exceeds the full corpus on 8 of 9 benchmarks; on Elliot-Pretrain, adding
Landsat-8/9 and using DEM and WorldCover as decoder targets improves most benchmarks, whereas giving those
maps to the encoder as inputs hurts. Finally, the proposed rule-based caption template matches small language model-generated captions.

Section~\ref{sec:background} reviews prior EO pretraining datasets.
Section~\ref{sec:methodology} describes how the dataset is built, its layers, its storage format and the
experimental setup. Section~\ref{sec:experiments} reports dataset statistics, the two pretraining
experiments and the caption study. Section~\ref{sec:discussion} discusses intended use and limitations,
Section~\ref{sec:conclusion} concludes, and Section~\ref{sec:data_access} explains how to access the data.
The Appendix gives further details on the construction, the metadata and the experiments.

\subsection{Related Work}
\label{sec:background}

Building ML-ready pretraining corpora from EO archives is non-trivial: each sensor
arrives in its own projection, resolution, and file format, and combining them into
aligned, batchable tensors requires substantial engineering effort. A growing set of
open datasets has taken on that work and their design choices (which sensors to
include, how to store them, and how to select locations) directly shape what a
model trained on them can learn. Table~\ref{tab:eo-datasets} summarises how current datasets compare.

\newcommand{\cmark}{\ding{51}}

% Present / absent / partial markers (xcolor is loaded by tikz)
\DeclareRobustCommand{\cmark}{\textcolor{green!45!black}{\ding{51}}}
\DeclareRobustCommand{\xmark}{\textcolor{black!25}{\ding{55}}}
% partial: the star is put in a zero-width box so the tick stays centred
\DeclareRobustCommand{\pmark}{\cmark\rlap{$^{\star}$}}
% rotated header, centred over its column (no parbox, so no offset)
\newcommand{\rot}[1]{\multicolumn{1}{c}{\rotatebox{90}{\textbf{#1}}}}

\begin{table}[h]
\caption{Comparison of large-scale EO pretraining datasets.
         \cmark\ present; \xmark\ absent; \cmark$^{\star}$ partial coverage;
         $^{\dagger}$ described for TerraMind \cite{jakubik2025terramind} but not part of the public TerraMesh release.}
\centering
\setlength{\tabcolsep}{3pt}
\renewcommand{\arraystretch}{1.2}
\begin{tabular}{%
  >{\raggedright\arraybackslash}p{3.1cm}%
  *{5}{>{\centering\arraybackslash}p{1.5cm}}%
}
\toprule
  & Major-TOM
  & SSL4EO-S12
  & Satlas Pretrain
  & Terra Mesh
  & Elliot-Pretrain (ours) \\
\midrule
\multicolumn{6}{l}{\textit{Modalities}} \\[2pt]
\quad S2 L2A          & \cmark & \cmark & \xmark & \cmark & \xmark \\
\quad S2 L1C          & \cmark & \cmark & \cmark & \cmark & \cmark \\
\quad S1 GRD          & \xmark & \cmark & \cmark & \cmark & \xmark \\
\quad S1 RTC          & \pmark & \xmark & \xmark & \cmark & \cmark \\
\quad Landsat (L8/9)  & \xmark & \xmark & \cmark & \xmark & \cmark \\
\quad NAIP            & \xmark & \xmark & \cmark & \xmark & \xmark \\
\quad DEM             & \pmark & \xmark & \xmark & \cmark & \cmark \\
\quad NDVI            & \xmark & \xmark & \xmark & \cmark & \xmark \\
\mbox{\quad LULC/WorldCover} & \xmark & \xmark & \xmark & \cmark & \cmark \\
\quad OSM context     & \xmark & \xmark & \xmark & \xmark & \cmark \\
\quad Text captions   & \xmark & \xmark & \xmark & \cmark$^{\dagger}$ & \cmark \\
\quad Text Tasks      & \xmark & \xmark & \xmark & \xmark & \cmark \\
\midrule
\multicolumn{6}{l}{\textit{Scale}} \\[2pt]
\quad Tiles / locations & 2.2M   & 251k     & 856k   & ${\sim}$9M & 279k    \\
\quad Storage           & 63\,TB & 1.5\,TB  & > 30\,TB     & 16\,TB     & 6.3\,TB \\
\midrule
\multicolumn{6}{l}{\textit{Storage \& access}} \\[2pt]
\quad Format          & GeoTIFF in Parquet & GeoTIFF (no index) & PNG tiles, 7 label types
                      & Zarr/ xarray & Parquet + COGs \\
\midrule
\multicolumn{6}{l}{\textit{Temporal structure}} \\[2pt]
\quad Mode            & Single & 4 seasons & Time series & Matched pair
                      & Mono / Monthly / Burst \\
\bottomrule
\end{tabular}
\label{tab:eo-datasets}
\end{table}

In \citet{francis2024majortom}, the authors released Major-TOM, one of the largest open
ML-ready Sentinel-2 collections: 2.2M patches on a uniform 10\,km grid, with
partial Sentinel-1 RTC and Copernicus DEM companions, totalling more than 60\,TB.
Its bottleneck is I/O, not spatial coverage: each spectral band is stored as an independent
GeoTIFF blob in its own Parquet column with one row per row group. SatlasPretrain \cite{bastani2023satlas} is widely used and genuinely effective, spanning Sentinel-2, Sentinel-1, Landsat 8/9, and NAIP imagery with 302M labels across 137 categories. Those labels, however, are split across seven different per-task formats, so using it as an unlabeled multi-sensor corpus means reconciling seven representations before a single aligned batch can be built.

SSL4EO-S12 \citep{wang2023ssl4eo} and TerraMesh \citep{blumenstiel2025terramesh}
target multimodal and multitemporal self-supervised pretraining. SSL4EO-S12 samples
251,079 locations worldwide and provides four seasonal acquisitions of co-located
Sentinel-1 GRD, Sentinel-2 L1C, and Sentinel-2 L2A imagery, distributed as plain
per-product GeoTIFF files with no unified index. TerraMesh is substantially larger, merging SSL4EO-S12 v1.1 with a filtered 
subset of Major-TOM into approximately 9 million co-registered samples 
across nine modalities (Sentinel-1 GRD/RTC, Sentinel-2 L1C/L2A, NDVI, 
Copernicus DEM, ESRI LULC, and natural-language captions created by LLaVA-Next \cite{li2024llava} and Overture Maps data \cite{marimo2025beyond}). It applies one temporal strategy uniformly across all locations, 
with S1/S2 pairs matched to within roughly one week of each other. It is the pretraining corpus behind TerraMind \cite{jakubik2025terramind}. PhilEO \citep{philEO2025scaling}
pretrains on the full 23\,TB Major-TOM Core-S2L2A, on FastTOM, a 2\,TB land-only subset of it,
and on the 0.5\,TB PhilEO Globe. The largest corpus helps only when model size grows with it,
and the seasonally structured PhilEO Globe, with four acquisitions per location, outperforms both
Major-TOM and FastTOM on land-cover mapping. This suggests that temporal structure can matter more
than raw scale, and that the benefit of scale depends on model size.

On the language side, SkyScript \citep{wang2024skyscript} turns 2.6M OpenStreetMap (OSM) \cite{openstreetmap}
tags into rule-based captions, and OSMDA \citep{ailuro2026osmda} prompts a VLM to caption
OSM-rendered map tiles, alongside broader web-scraped RS caption datasets
(RS5M/GeoRSCLIP, GAIA). These are flat image-text collections distributed
independently of any multi-sensor pretraining corpus. TerraMesh takes a step toward
closing this gap by including captions \citep{blumenstiel2025terramesh}, but its temporal
structure is uniform across all locations and its Zarr-based storage couples pixel retrieval to full archive loads, preventing the metadata-first filtering that TACO's separated Parquet index enables. MajorTOM:Elliot Pretrain instead generates language from the dataset's own metadata, across heterogeneous temporal regimes and with facts tagged by modality, so that captions and tasks follow any multi-sensor batch without coordinate reconciliation: a layout optimized for GFM adaptation to VLMs.

\section{Materials and Methods}
\label{sec:methodology}
\subsection{Dataset Overview}
\label{sec:overview}

MajorTOM:Elliot Pretrain is a multi-sensor, multi-temporal Earth observation
pretraining dataset built on the Major-TOM 10\,km global grid
\citep{francis2024majortom}, distributed through Source Cooperative at
\texttt{source.coop/major-tom/elliot-pretrain} under CC-BY-SA 4.0 (except its OpenStreetMap files and the \texttt{facts/} folder, which are ODbL-1.0) (see Fig.\ref{fig:graphical-abstract}). It is a single collection of three TACO v3 subdatasets (Monotemporal, Monthly and Burst). Each tile folder holds its sensor, cloud-mask, OpenStreetMap and administrative layers, and per-image metadata such as ERA5 weather. A top-level \verb|facts/| folder of precomputed fact sheets lets captions and tasks be generated without reading pixels. The three subdatasets
share a single tile identifier space and a common $1056\times1056$-pixel, 10\,m
spatial grid.

The 279{,}166 tiles are drawn from the \textit{land grid}: the $\approx$2.5 million global-grid locations where Sentinel-2 imagery exists, selected for semantic diversity as described in Section~\ref{sec:elliot-collection}.

\begin{figure}[h]
\centering
\includegraphics[width=\linewidth]{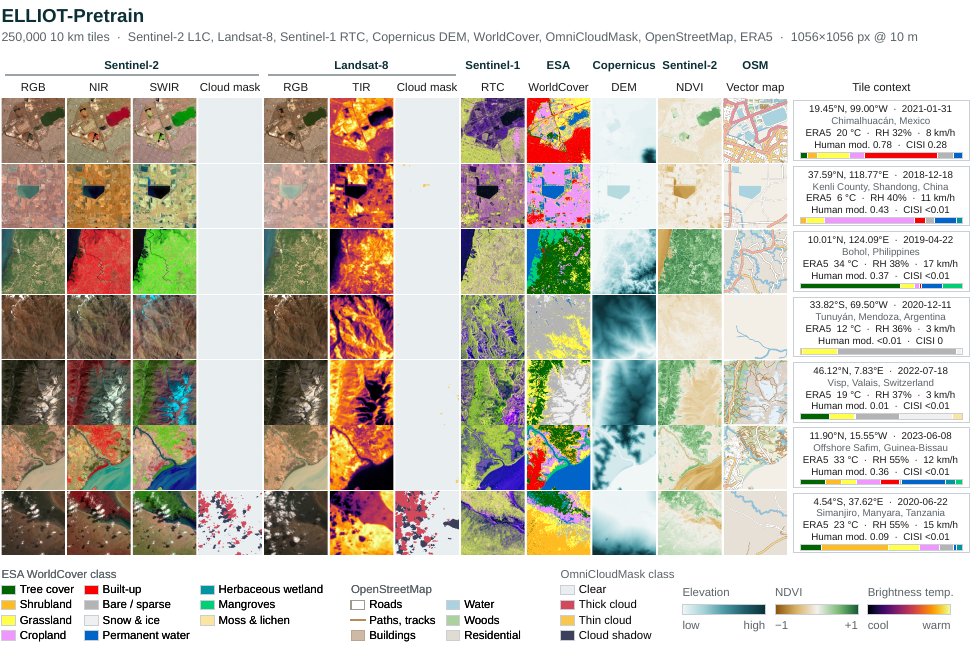}
    \caption{Example tiles across seven landscape types (one per row), with all co-registered layers and the per-tile context.
\emph{Sentinel-2 L1C:} RGB, true colour (bands B4, B3, B2); NIR, false-colour near-infrared (B8, B4, B3), in which vegetation appears red; SWIR, false-colour shortwave infrared (B12, B8, B4); cloud mask, OmniCloudMask classes.
\emph{Landsat-8:} RGB, true colour (B4, B3, B2); TIR, thermal infrared (band 10), a brightness temperature; cloud mask as above.
\emph{Sentinel-1 RTC:} radiometrically terrain-corrected backscatter in dB, with VV (red), VH (green) and VV$-$VH (blue) as radar polarisations.
\emph{Other layers:} WorldCover, ESA land-cover map; DEM, Copernicus GLO-30 digital elevation model; NDVI, normalised difference vegetation index of Sentinel-2, $(\mathrm{B8}-\mathrm{B4})/(\mathrm{B8}+\mathrm{B4})$; vector map, OpenStreetMap features of the tile. Thermal, radar and elevation colours are stretched per tile, and class colours are given in the legends.
\emph{Tile context:} tile-centre latitude and longitude, Sentinel-2 acquisition date and place; ERA5 (ECMWF reanalysis) 2\,m air temperature, relative humidity (RH) and 10\,m wind speed at that date; Global Human Modification index (Human mod., 0 to 1); Critical Infrastructure Spatial Index (CISI); and the share of each WorldCover class in the tile (coloured bar).}

    \label{fig:diversity}
\end{figure}

This dataset holds five modalities: Sentinel-2 L1C \cite{drusch2012sentinel},
Landsat-8/9 OLI-TIRS \cite{roy2014landsat}, Copernicus DEM GLO-30 \citep{copdem2021}, ESA WorldCover v200 \cite{zanaga2022esa} and Sentinel-1 RTC (VV and
VH) \cite{torres2012gmes} across the selected 279{,}166 tiles (see Fig. \ref{fig:diversity}).

\begin{table}[h]
\centering
\caption{The three subdatasets, one per temporal regime. Each of the 279{,}166 selected tiles belongs to exactly one
regime. Acquisitions are per sensor.}
\label{tab:regimes}
\small
\begin{tabular}{lrrll}
    \toprule
    Regime & Tiles & Share & Acquisitions per tile & Selection of dates \\
    \midrule
    Monotemporal & 250{,}000 & 89.6\% & 1 & least cloudy available \\
    Monthly & 12{,}500 & 4.5\% & 12 & greedy, maximal circular day-of-year spread \\
    Burst & 16{,}666 & 6.0\% & 6 & greedy spread within the best 30-day window \\
    \midrule
    Total & 279{,}166 & 100\% & & \\
    \bottomrule
\end{tabular}
\end{table}

\subsection{Tile selection and temporal regimes}
\label{sec:elliot-collection}

The \textit{land grid}: $\approx$2.5 million
tiles of the global 10\,km grid for which Sentinel-2 imagery exists, is
obtained by retaining only global-grid tiles that returned at least one valid
acquisition. This is the candidate pool, i.e.\ every location
that could in principle be included, not the released collection. The land grid
carries only scalar metadata (identifier, CRS, centroid, grouping codes) and no
pixels; it is defined by Sentinel-2's orbital acquisition plan rather than an
external land mask.

From this pool, 279{,}166 tiles are selected for semantic
diversity rather than geographic coverage: two tiles count as similar only
if a vision foundation model's embeddings of their imagery are similar, independent
of ground distance. More specifically, every land-grid tile is embedded with AlphaEarth Foundations \citep{brown2025alphaearth}, chosen because it is a state-of-the-art GFM with publicly available embeddings (64-d, spatially averaged and renormalised to unit norm), and a two-level spherical
$k$-means hierarchy over the $\approx$2.5M embeddings selects 279{,}840
representative tiles, one per fine cluster, in minutes rather than the hours a
flat $k$-means at that scale would require (Appendix~\ref{sec:supplement_collection}). 
Intersecting these candidates against tiles present in both the Sentinel-2 and
Landsat-8/9 tables removes 674 high-latitude tiles lacking L8 coverage, leaving
279{,}166 tiles that are randomly split with a fixed seed into Monotemporal,
Monthly, and Burst at the proportions in Table~\ref{tab:regimes}.
Monotemporal receives by far the largest share of the selected tiles because each additional acquisition of the same location is highly redundant with the previous ones, so temporal tiles add little new signal per byte and per training step, whereas Monthly and Burst are kept at a smaller size because many downstream tasks require temporal inputs and the model must learn what time means.

Temporal selection differs by regime. Monotemporal reuses each tile's Major-TOM 
image identifier if one already exists, for compatibility with prior
pretraining runs, and otherwise takes the least-cloudy available acquisition.
Sentinel-2 and Landsat-8/9 dates are chosen independently with no cross-sensor
alignment enforced. Monthly and Burst are both driven by Sentinel-2 acquisition
dates, chosen greedily to maximise temporal spread: circularly across the full
year for Monthly's 12 slots and linearly within an automatically selected 30-day
window for Burst's 6, with Landsat-8/9 matched to each date through a four-level
cascade of progressively relaxed temporal tolerance and cloud-free score threshold
(Appendix~\ref{sec:supplement_collection}). A small number of tiles have no usable Landsat-8/9
acquisition: 26 of 12{,}500 Monthly tiles (all twelve images, 23 of them south of 60\textdegree{} S, where
WRS-2 is intrinsically gapped), 18 of 250{,}000 Monotemporal tiles and 3 of 16{,}666 Burst tiles (13 images). These are marked as nodata. 

\subsection{Layers}
\label{sec:sensors}

Every tile holds the same set of layers on one shared $1056\times1056$-pixel, 10\,m grid: four sensor
images (Sentinel-2, Landsat-8/9, Sentinel-1, and the static DEM and WorldCover maps), and context
layers (cloud masks, ERA5 weather and OpenStreetMap) that are not sensor images but share the tile's
grid and folder. Two rules apply to all sensor layers. First, each sensor fills the acquisition slots
of the tile's regime (one in Monotemporal, 12 in Monthly, 6 in Burst) with a single image per slot,
typically the least cloudy one available. Second, if a sensor has no acquisition for a slot, the
image is still stored, filled with the nodata value, so that every tile has the same layout.

\paragraph{Optical: Sentinel-2 and Landsat-8/9.} Sentinel-2 L1C has bands at three native resolutions
(10, 20 and 60\,m). The 20 and 60\,m bands are brought to 10\,m by integer pixel replication, without
interpolation, and all 13 bands are stored in one COG, scaled by $10{,}000$ following the ESA convention.
Landsat-8/9 combines 9 optical (OLI) and 2 thermal (TIRS) bands in one 11-band COG; because the two
groups have different noise floors, they receive different bit-depth trims
(Appendix~\ref{sec:supplement_sensors}). 807 Monotemporal Landsat frames are OLI-only products whose
thermal bands 10 and 11 are zero. Landsat-8/9 gaps are more common than Sentinel-2 gaps, mainly at high
latitude, where the WRS-2 grid leaves areas that Sentinel-2's wider swath covers.

\paragraph{Radar: Sentinel-1.} Sentinel-1 RTC is stored as two bands, VV and VH, in linear power
($\gamma^0$, float32, mantissa truncated to 12 bits), with one image per Sentinel-2 acquisition. For each
Sentinel-2 image the VV+VH pass closest in time is selected, without temporal interpolation; HH+HV passes
are excluded. The source, the Planetary Computer \texttt{sentinel-1-rtc} product (CC-BY-4.0), is already
on the tile grid, so an image is resampled only if the UTM zone of the pass differs from the tile's.
A residual offset to Sentinel-2 of about 1 to 1.5 pixels remains and is not corrected
(Appendices~\ref{sec:supplement_auxiliary} and~\ref{sec:validation}). If no VV+VH pass exists, mainly
in polar regions, where Sentinel-1 acquires HH+HV, and over open sea, the image is stored as nodata (0).
Over the three regimes, 93.2\% of samples have valid Sentinel-1 in at least 90\% of the pixels of every
image, and 6.6\% have none. Per-sample fields (\texttt{s1:valid\_fraction}, \texttt{s1:valid\_fraction\_min},
\texttt{s1:frames\_with\_data}) allow filtering, and per-image fields record the pass date, its gap to the
Sentinel-2 date and its orbit (Table~\ref{tab:schema}).

\paragraph{Static layers: DEM and WorldCover.} Copernicus DEM GLO-30 is the only layer that is genuinely
interpolated: it is resampled bilinearly from its native 30\,m grid to 10\,m. ESA WorldCover is
categorical, so horizontal differencing is disabled and no bit-depth trim is applied. Both layers are
stored once per tile, since they do not change over time.

\paragraph{Cloud masks and weather.} Every Sentinel-2 and Landsat-8/9 image has an OmniCloudMask cloud
and shadow mask (clear, thick cloud, thin cloud, shadow), stored as a 10\,m PNG, with its class fractions
in the per-image tables. Hourly ERA5 ~\cite{soci2024era5}  meteorology (2\,m temperature and dew point, humidity, precipitation, cloud
cover and 10\,m wind, at a spatial resolution of $0.25^\circ$) is read at the hour nearest each acquisition and stored as
columns of the per-image tables (\texttt{ml:era5\_*}), so it has no raster of its own.

\paragraph{OpenStreetMap.} OpenStreetMap is stored as vector data. Each tile has an \texttt{osm.parquet}
of features (75.5 million entities in total) and an \texttt{admin.parquet} of administrative units
(1.55 million in total), both GeoParquet files in the tile's UTM CRS. Each feature has its OSM identifier,
category, name and local name, area or length clipped to the tile, Wikidata and Wikipedia links and its
full tags (Table~\ref{tab:schema}), and per-tile counts are kept in the sample table. The OSM data is
used in two steps. The extraction keeps all features above a small size limit, so that tasks such as
small-object detection can use ordinary features like a restaurant or a bus stop. When the task library
builds a tile's fact sheet, a second, stricter selection keeps only the named features a description
should mention (Appendix~\ref{sec:supplement_auxiliary}).

\paragraph{Text.} The collection does not store captions. It stores the data needed to produce them,
and a code library (\texttt{majortom-elliot-tasks}, MIT licence~\cite{majortom_elliot_tasks}) builds
text from the layers when it is needed, so the wording changes at every epoch. For each tile the library
assembles a fact sheet of everything measured: spectral indicators from Sentinel-2 and Landsat-8/9,
relief and land cover, named OpenStreetMap features, cloud fractions, ERA5 weather and Sentinel-1
backscatter, plus time-series facts for Monthly and Burst tiles. Every fact is tagged with the sensor
that supports it, so facts from a sensor the model is not given are dropped. From the sheet the library
generates 16 families of vision-language tasks (captioning, grounding, grids, numbers and time-series
questions), each scored from 0 to 1 by a rule for its answer shape; captions are written by a rule-based
template that uses no language model. Appendix~\ref{sec:language} lists the task families and describes
the scoring and the template.

\subsection{Storage format and access}
\label{sec:design-rationale}

MajorTOM:Elliot Pretrain is built around three design choices that apply uniformly
across all subdatasets and that together determine how training batches are
constructed.

\paragraph{Layer modularity.} Every sensor and context layer is stored in its own subfolder of the tile
(\texttt{DATA/<tile>/s2/}, \texttt{l8/}, \texttt{s1/}, \texttt{dem/}, \texttt{lc/}, \texttt{cloud\_s2/},
\texttt{cloud\_l8/}), with its own \texttt{children\_\_<layer>.parquet} metadata table. A tile is
identified by its folder index within a regime, and \texttt{majortom:code\_10km} gives its grid code. Because every
COG is already aligned to the same geotransform, layers combine into pixel-aligned tensors with no reprojection.
A user downloads only the regimes and layers they need, and a new layer can be
added as a further subfolder and metadata table alongside the existing ones.

\paragraph{Format and access.} Pixel payloads are stored as Cloud-Optimised
GeoTIFFs (COGs), which support HTTP range requests so that a reader retrieves
only the bytes it needs. All COGs are resampled to a shared $1056\times1056$-pixel, 10\,m grid, whose side is divisible
by the common ViT patch sizes and by the $3\times3$ internal tiling (Appendix~\ref{sec:supplement_design}).
Pixel payloads are
stored separately from the Parquet metadata, which follows the TACO~v3 layout,
so any Parquet-capable engine can filter on metadata, cloud cover, temporal
regime, etc.\ before a single pixel is read. The \texttt{taco.ml} reader decodes
each sample into arrays on demand. Internally, COGs use ZSTD compression with
$352\times352$-pixel tiles, sized for GPU-native decompression via nvCOMP. Full
compression parameters per sensor are given in
Section~\ref{sec:supplement_sensors}. 

\paragraph{Temporal and contextual design.} MajorTOM:Elliot Pretrain partitions
its 279{,}166 locations into three sub-datasets targeting different training
signals: Monotemporal for broad static pretraining, Monthly for seasonal and
phenological dynamics, and Burst for short-timescale events
(see Table~\ref{tab:regimes}). Each location appears in
exactly one regime, so the three sub-datasets together define a training
curriculum without overlap. Each tile also carries the context layers Section~\ref{sec:sensors}. A task library
turns these row-aligned layers into fact sheets and vision-language tasks at load time, making
language-grounded pretraining a native feature of the dataset rather than a separately
distributed resource.

\subsubsection{File Layout and Size}
\label{sec:layout}

MajorTOM:Elliot Pretrain is organized into three regimes (Table~\ref{tab:layout}). In each one a tile is a folder \texttt{DATA/<tile>/}, named by an integer index, whose subfolders hold one file per acquisition: \texttt{data.tif} in Monotemporal and
\texttt{img\_NN.tif} (Monthly, Burst) for the rasters, with the elevation and land-cover layers stored once as
\texttt{main.tif}.  Cloud masks are PNGs, sensors are stored as rasters, and each tile has two OSM tables. The collection occupies 6.27\,TB (3.08\,TB Monotemporal, 1.84\,TB the Monthly and 1.35\,TB the Burst) plus the \texttt{facts/} folder.
A Sentinel-2 image takes 5.7\,MB on average (against 29\,MB uncompressed, $5\times$), Landsat-8/9 2.3\,MB
($21\times$), Sentinel-1 4.1\,MB, the elevation and land-cover rasters 73 and 44\,KB, and a cloud mask 3--8\,KB.
The per-tile OSM and administrative tables average 77 and 14\,KB. Averages are over 300 random Monotemporal tiles.
The metadata tables are plain Parquet files with 10{,}000 rows per row group; the join key is
\texttt{internal:parent\_id}: image rows point to their layer-folder row in \texttt{children.parquet}, whose \texttt{internal:parent\_id} is the tile.

\begin{table}[tbp]
\centering
\caption{Directory layout of the collection. The three sub-datasets \texttt{monotemporal/}, \texttt{monthly/} and
\texttt{burst/} share it; only the file names inside a tile's layer folders differ. Monotemporal stores one
acquisition per layer, Monthly and Burst one file per acquisition, with \texttt{NN} running over 00--11 (Monthly)
or 00--05 (Burst). Static layers (DEM, land cover) are stored once.}
\label{tab:layout}
\small
\renewcommand{\arraystretch}{1.15}
\begin{tabular}{
  >{\raggedright\arraybackslash}p{5.1cm}
  >{\raggedright\arraybackslash}p{2.1cm}
  >{\raggedright\arraybackslash}p{2.3cm}
  >{\raggedright\arraybackslash}p{4.6cm}}
\toprule
\textbf{Path} & \textbf{Monotemporal} & \textbf{Monthly / Burst} & \textbf{Content} \\
\midrule
\pth{<regime>/} & & & \\
\addlinespace[2pt]
\pth[1.2]{COLLECTION.json}
  & \multicolumn{3}{>{\raggedright\arraybackslash}p{9.6cm}}{TACO collection description, column descriptions, band and radiometry contract} \\
\addlinespace[6pt]
\pth[1.2]{METADATA/} & & & \\
\addlinespace[2pt]
\pth[2.4]{sample.parquet}
  & \multicolumn{3}{>{\raggedright\arraybackslash}p{9.6cm}}{One row per tile: grid codes, tile-level context, geometry, OSM counts,
    WorldCover and DEM statistics, cloud and Sentinel-1 summaries} \\
\addlinespace[4pt]
\pth[2.4]{children.parquet}
  & \multicolumn{3}{>{\raggedright\arraybackslash}p{9.6cm}}{One row per top-level entry of a tile (layer folders, \texttt{osm.parquet}, \texttt{admin.parquet})} \\
\addlinespace[4pt]
\pth[2.4]{children\_\_<layer>.parquet}
  & \multicolumn{3}{>{\raggedright\arraybackslash}p{9.6cm}}{One row per file (s2, l8, s1, cloud\_s2, cloud\_l8, dem, lc): dates, product, scores,
    geometry, ERA5, mask class fractions, Sentinel-1 pass} \\
\addlinespace[6pt]
\pth[1.2]{DATA/<tile>/} & & & \\
\addlinespace[2pt]
\pth[2.4]{s2/}, \texttt{l8/}
  & \texttt{data.tif} & \texttt{img\_NN.tif} & Sentinel-2 L1C (13 bands), Landsat-8/9 (11 bands) \\
\addlinespace[4pt]
\pth[2.4]{s1/}
  & \texttt{data.tif} & \texttt{img\_NN.tif} & Sentinel-1 RTC, VV and VH \\
\addlinespace[4pt]
\pth[2.4]{cloud\_s2/}, \texttt{cloud\_l8/}
  & \texttt{data.png} & \texttt{img\_NN.png} & Cloud and shadow masks \\
\addlinespace[4pt]
\pth[2.4]{dem/}, \texttt{lc/}
  & \texttt{data.tif} & \texttt{main.tif} & DEM and WorldCover (1 band, static) \\
\addlinespace[4pt]
\pth[2.4]{osm.parquet}, \texttt{admin.parquet}
  & \multicolumn{3}{>{\raggedright\arraybackslash}p{9.6cm}}{OpenStreetMap entities and administrative units} \\
\midrule
\pth{facts/<regime>/} & & & \\
\addlinespace[2pt]
\pth[1.2]{facts-NNNNNN.parquet}
  & \multicolumn{3}{>{\raggedright\arraybackslash}p{9.6cm}}{Precomputed per-tile facts (plain Parquet, ODbL-1.0)} \\
\bottomrule
\end{tabular}
\end{table}

\subsection{Experimental Setup}
\label{sec:exp-setup}

\paragraph{Data.} Experiment~1 pretrains on Core-S2L2A, the Major-TOM Sentinel-2 L2A collection (2{,}184{,}259 tiles after filtering), and Experiment~2 on Elliot-Pretrain (279{,}166 tiles:
250{,}000 monotemporal, 12{,}500 monthly and 16{,}666 burst), whose Sentinel-2 data is L1C.
Each corpus is normalized with its own per-band statistics.

\paragraph{Pretraining.} We choose OlmoEarth, one of the best EO encoders in the literature, with open weights, recipes and code, available in several model sizes. We pretrain OlmoEarth Base from scratch with the official v1.2 recipe \citep{olmoearth2025,olmoearth2026v12}:
latent masked image modelling with a patch-discrimination loss and an InfoNCE term, AdamW (learning rate $10^{-4}$, weight decay 0.02),
global batch 512, token budget 2250, a cosine schedule with warmup over 1.2\% of the run, and fixed seed. The model has 88.8M parameters with Sentinel-2 as its only modality; the Experiment~2 models, which add modalities as encoder inputs or decoder targets, have 92.0M to 101.4M (Table~\ref{tab:app-hparams}). Only the data, the
modalities, the temporal partition and the number of steps change between runs. One epoch is 488 steps over the 250{,}000 monotemporal
tiles and 4{,}266 steps over Core-S2L2A, which represent 0.07\% and 0.64\% of the official 667{,}200 steps. We use short schedules to run many
experiments at low cost, and conclusions are limited to this regime. The ten-epoch runs of Appendix~\ref{sec:exp-base} (42{,}660 steps) reach 6.4\% of the official training regime.

\paragraph{Evaluation.} After pretraining, the encoder is frozen. Each benchmark input is standardized with the benchmark's own per-band
mean and standard deviation. On the pooled features we fit a $k$-nearest-neighbour (KNN) classifier (EuroSAT, So2Sat, BrickKiln, BigEarthNet, TreeSatAI)
or a linear probe (segmentation and regression), and report the test score of the probe after its 50 training epochs (the validation score is computed once, at the end, so no epoch is selected; the KNN classifier has no epochs). The metrics are accuracy
(EuroSAT, So2Sat, BrickKiln), micro-F1 (BigEarthNet), macro-F1 (TreeSatAI), mIoU (PASTIS, MADOS) and RMSE (Open-Canopy in m,
BioMassters in Mg\,ha$^{-1}$, lower is better). We use nine benchmarks to evaluate Experiment~1 (Section~\ref{sec:exp-base}, Fig.~\ref{fig:trajectory}) and Experiment~2 (Section~\ref{sec:exp-modality}, Fig.~\ref{fig:exp2-heat}). Models with a Sentinel-1 pathway are scored on the Sentinel-1+Sentinel-2 version of the benchmark if one exists. Each configuration is pretrained once. For the ten-epoch runs, each checkpoint is evaluated with three probe seeds (3622, 8956, 30991) and averaged. This measures probe noise only, because the encoder is fixed. Full details are in Appendix~\ref{sec:app-config} and Appendix~\ref{sec:app-exp2}.

\paragraph{Captions.} The library writes captions with a template (Section~\ref{sec:sensors}; examples in Section~\ref{sec:supplement_captions}). We ask how good they are and what a language
model would add. We compare four ways of writing a caption from the same fact sheet. One uses a \emph{template}, a program with fixed rules and no language
model. Three use a language model (LLM) that receives
the fact sheet: GPT-6-luna (small), GPT-6.1-sol (large), and GPT-6.1-sol with the image in addition.

\section{Results}
\label{sec:experiments}

This section first describes the released collection through its statistics (Section~\ref{sec:stats}): tile counts and geographic spread, acquisition dates, cloud cover, Sentinel-1 coverage and OpenStreetMap context. It then uses the collection for two pretraining experiments and a caption study. Experiment~1 asks how the size and curation of the pretraining data affect downstream accuracy (Appendix~\ref{sec:exp-base}). Experiment~2 asks whether additional sensors (part a) and the Monthly and Burst temporal partitions (part b) help (Appendix~\ref{sec:exp-modality}). The caption study compares template captions with captions written by language models (Appendix~\ref{sec:exp-captions}). The pretraining experiments use short schedules and one run per configuration, and evaluate frozen encoders on nine downstream benchmarks (Section~\ref{sec:exp-setup}).

\subsection{Dataset Statistics}
\label{sec:stats}

\begin{figure}
    \centering
    \includegraphics[width=\linewidth]{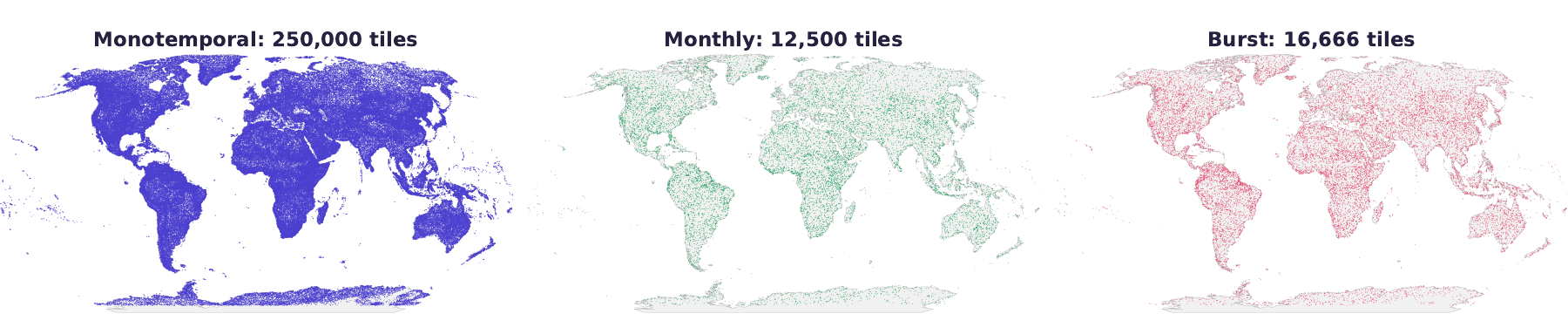}
    \caption{Tile locations spread across the earth for the three temporal regimes}
    \label{fig:stats}
\end{figure}

Fig.~\ref{fig:stats}, Tab.~\ref{tab:regime-stats} and the text below summarise the released collection, computed from its TACO metadata.
The collection holds 279{,}166 tiles and 499{,}996 Sentinel-2 images in the three regimes together
(250{,}000 + 150{,}000 + 99{,}996), with one Landsat-8/9 and one Sentinel-1 image per
Sentinel-2 image. Tiles are spread over all land masses (Fig.~\ref{fig:stats}; of the Monotemporal tiles, 33.4\% lie between
30\textdegree{} and 60\textdegree{} N, 27.0\% between the equator and 30\textdegree{} N, 20.1\% between
30\textdegree{} S and the equator, 3.9\% between 60\textdegree{} S and 30\textdegree{} S, and 15.6\% poleward of 60\textdegree{} in either hemisphere, which
includes Antarctica and Greenland.

\paragraph{Acquisition dates.} Monotemporal images are dominated by 2019--2024 (93\% of the images), because
the single acquisition is the Major-TOM image if one exists and the least cloudy available
otherwise. Monthly spans 2015--2026 and Burst 2017--2025, since the date selection of both regimes is
free to pick any year of the archive (Section~\ref{sec:elliot-collection}).  The Monthly regime covers the annual cycle. The median over tiles of the smallest
circular day-of-year gap between two selected acquisitions is 22\,days, against 30.4\,days for twelve
perfectly even dates, and the median largest gap is 46\,days. For 843 tiles (6.7\%) the largest gap
exceeds 130\,days, which happens where the archive has no usable acquisition for part of the year.

\paragraph{Cloud cover.} Monotemporal images are selected to be clear: by the CloudScore+ clear score, 100\% of the Sentinel-2 images have a cloudiness (one minus the score) below 0.1, and by the OmniCloudMask masks 90.3\% have a cloud fraction below 0.1. Monthly and Burst keep cloudy acquisitions on purpose, because the selection maximises temporal
spread rather than clarity (Tab.~\ref{tab:regime-stats}): the 90th percentile of the cloudiness (one minus the CloudScore+ clear score) of the Sentinel-2 images is
0.87 for Monthly and 0.28 for Burst. The per-pixel four-class masks allow
users to filter or weight these acquisitions.

\paragraph{Sentinel-1.} In all three regimes 93.2\% of the samples have valid Sentinel-1 in at least 90\% of
the pixels of every image, and 6.6\% have none. The median gap between a Sentinel-1 pass and its
Sentinel-2 image is 2.4\,days in Monotemporal, 2.4\,days in Monthly and 2.0\,days in Burst, and 58\%, 57\%
and 65\% of the valid images lie within three days. The tail differs between regimes: the 90th percentile is
5.9\,days in Burst and 39\,days in Monthly, but 162\,days in Monotemporal, so users who need close
temporal alignment should filter on \texttt{s1:gap\_days}.

\paragraph{OpenStreetMap context.} Over the three regimes the collection holds 75.5\,million OSM entities
(67.8M, 3.4M and 4.3M in Monotemporal, Monthly and Burst; 23.3M of the Monotemporal entities carry a name)
and 1.55\,million administrative units. 82\% of the tiles contain at least one named entity, with a median of
three per tile.

\begin{table}[h]
\centering
\small
\caption{Statistics of the released regimes. Cloudiness bins are the share of images with cloudiness (1 $-$ clear score) in $<$0.1 / 0.1--0.3 / 0.3--0.7 / 0.7--0.9 / $>$0.9.}
\begin{tabular}{lccc}
\toprule
 & Monotemporal & Monthly & Burst \\
\midrule
Tiles                                          & 250{,}000          & 12{,}500             & 16{,}666 \\
Images per sensor                              & 250{,}000          & 150{,}000            & 99{,}996 \\
Sentinel-2 acquisition years                   & 2015--2026         & 2015--2026           & 2017--2025 \\
Cloud fraction (masks), median                 & 0.00               & 0.19                 & 0.07 \\
Sentinel-2 cloudiness bins (\%)                & 100 / 0 / 0 / 0 / 0 & 74 / 6 / 8 / 3 / 10 & 85 / 5 / 5 / 1 / 3 \\
Landsat-8 cloudiness bins (\%)                 & 94 / 3 / 3 / 0 / 0 & 69 / 8 / 10 / 4 / 9  & 75 / 11 / 11 / 3 / 0 \\
Landsat-8 images without data                  & 18                 & 312                  & 13 \\
Sentinel-1 valid ($\geq$90\% pixels)           & 93.2\%             & 93.1\%               & 93.2\% \\
Sentinel-1 gap, median / P90 (days)            & 2.4 / 162          & 2.4 / 39             & 2.0 / 5.9 \\
OSM entities (M)                               & 67.8               & 3.4                  & 4.3 \\
Administrative units (M)                       & 1.39               & 0.07                 & 0.09 \\
Tiles poleward of $60^\circ$                   & 15.6\%             & 15.9\%               & 15.4\% \\
\bottomrule
\end{tabular}
\label{tab:regime-stats}
\end{table}

\subsection{Experiment 1: Pretraining data size and curation}
\label{sec:exp-base}

\begin{figure}[htb]
\centering
\includegraphics[width=\linewidth]{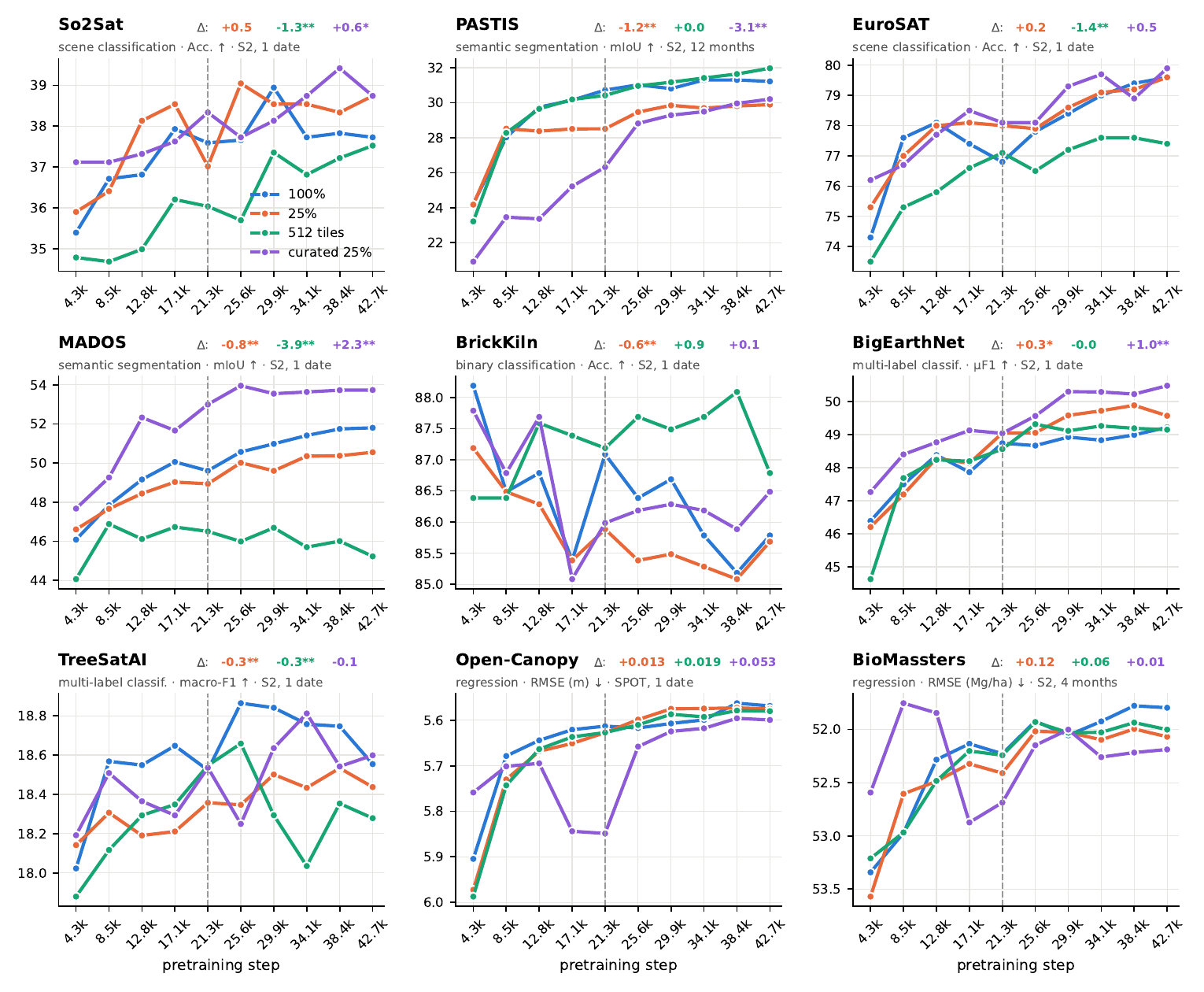}
\caption{OlmoEarth Base on Core-S2L2A: downstream scores at matched optimizer steps for the
full corpus (100\%), a random 25\% subset, a curated 25\% subset and 512 random tiles, all trained for ten epochs (42{,}660 steps) in two parts; the dashed line marks step 21{,}330, where training was resumed with a restarted learning-rate schedule.
The vertical axes of the two RMSE
panels (Open-Canopy, BioMassters) are inverted so that up is better in every panel.}
\label{fig:trajectory}
\end{figure}

We ask whether more data, or better-curated data, helps to pretrain a GFM under the same pretraining regime. The curated subset is selected on the same principle as Elliot-Pretrain, hierarchical $k$-means clustering of AlphaEarth embeddings (Sections~\ref{sec:elliot-collection} and~\ref{sec:app-config}), so this experiment is the evidence for that design choice, tested on a corpus whose full, unselected pool is available for comparison. We compare four sets drawn from Core-S2L2A: the full corpus, a random 25\% subset, the AlphaEarth-curated 25\% subset and 512 random tiles. Then we pretrain OlmoEarth for ten full-corpus epochs (42{,}660 steps) and evaluate every 4{,}266 steps. Training was done in two parts: five epochs (21{,}330 steps) with their own cosine schedule, then five more, resumed from the
step-21{,}330 checkpoint with the schedule recomputed over all 42{,}660 steps. Therefore, at step 21.3K the learning rate jumps back up to the midpoint of a cosine and decays again. Figure~\ref{fig:trajectory} shows the scores at every checkpoint. To support the comparisons we run a paired $t$-test over the ten checkpoints, comparing two runs at the same checkpoint, and report the mean difference with its 95\% confidence interval (CI) in Table~\ref{tab:trajectory-paired}; the number above each panel is this mean difference to the 100\% run, with $^*$ for $p<0.05$ and $^{**}$ for $p<0.01$.
The curated 25\% subset is significantly better than the full corpus on MADOS ($+2.32$ points, CI $[+1.78,+2.87]$), BigEarthNet ($+0.99$, $[+0.71,+1.28]$) and So2Sat ($+0.59$, $[+0.02,+1.17]$). It is indistinguishable from it (CI includes zero) on EuroSAT, BrickKiln, TreeSatAI, Open-Canopy and BioMassters, and worse only on PASTIS ($-3.15$, $[-4.46,-1.83]$). It is therefore competitive with or better than 100\% on 8 of 9 benchmarks. At the last checkpoint the extreme gaps are $+1.93$ in MADOS and $-1.01$ in PASTIS.
Compared directly with the random 25\% subset (last column of Table~\ref{tab:trajectory-paired}), the curated subset is better on MADOS ($+3.09$, $[+2.35,+3.83]$), BigEarthNet ($+0.68$), BrickKiln ($+0.62$) and TreeSatAI ($+0.13$), level on So2Sat, EuroSAT and the two regression tasks, and worse on PASTIS ($-1.98$, $[-3.46,-0.49]$). The advantage over random selection is real on four benchmarks but not uniform.
The 512-tile run is significantly worse than 100\% on MADOS ($-3.93$), EuroSAT ($-1.38$), So2Sat ($-1.30$) and TreeSatAI ($-0.33$), with a final-checkpoint gap of up to $6.57$ points in MADOS, but level on BigEarthNet ($-0.02$), PASTIS ($+0.04$) and BrickKiln ($+0.89$, $[-0.04,+1.83]$). Those tasks may be less sensitive to scale.
These intervals are over checkpoints of runs that were each pretrained once and no correction is made for multiple comparisons, so they describe the training trajectory and not the variation between pretraining seeds (Appendix~\ref{sec:app-exp1}). 

Regression tasks (Open-Canopy, BioMassters) show near-zero differences across all conditions (every CI includes zero), and training steps seem to matter much more than which tiles were used. We observe that with more data, the frozen features separate BrickKiln slightly worse as training goes on. With 512 tiles they get slightly better. The fall is mostly in the first 17k steps. The curated subset favors rare landscape types and drops many near duplicate typical scenes. PASTIS is crop parcels in France, so it may rely on exactly those typical agricultural scenes, which could explain a slower start on PASTIS, and the high performance of MADOS. Nevertheless, this experiment would benefit from a more expensive longer pretraining comparison as it is still about 6.4\% of the official OlmoEarth step count.

\subsection{Experiment 2: Sensors and temporal partitions}
\label{sec:exp-modality}

\begin{figure}[h]
\centering
\includegraphics[width=\linewidth]{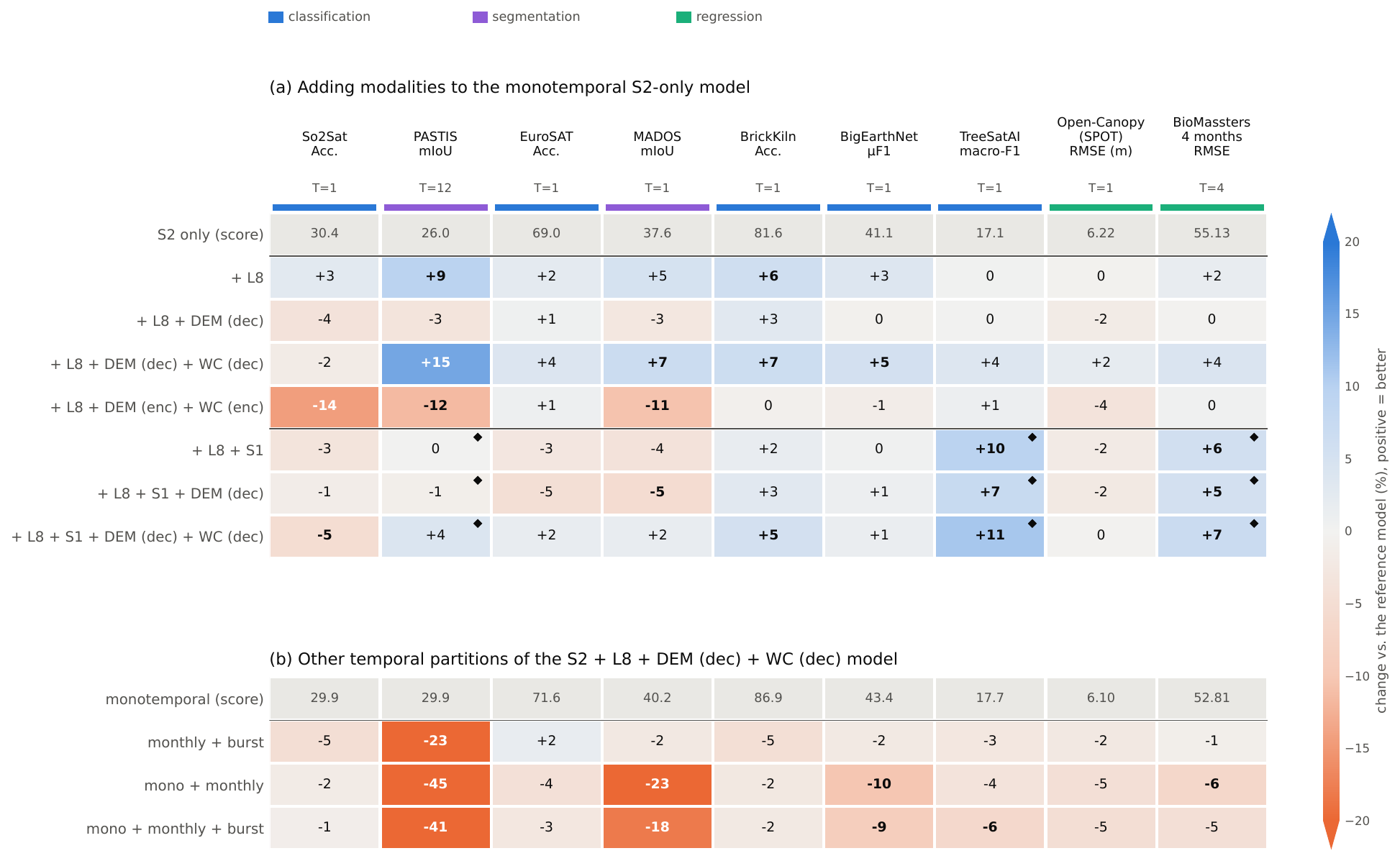}
\caption{Change in downstream score at step 488 (OlmoEarth Base, Elliot-Pretrain, frozen
encoder) relative to (a)~the S2-only model and (b)~the monotemporal S2+L8+DEM+WC model. Positive is
better; RMSE tasks are sign-flipped. Note that dec = decoder-only target: the encoder never sees it and the decoder predicts it from the visible Sentinel-2 and Landsat patches; enc = encoder input: DEM and WorldCover are given to the encoder like extra sensor bands. Under each column: number of time steps T given to the model.
Diamond = the model has a Sentinel-1 pathway and this cell is scored with Sentinel-1 + Sentinel-2 input (PASTIS, TreeSatAI, BioMassters); all other cells use optical input. The reference is the S2-only model in (a) and the monotemporal S2+L8+DEM(dec)+WC(dec) model in (b), each scored on its own optical input.}
\label{fig:exp2-heat}
\end{figure}

We ask how useful additional sensors and time series are across diverse benchmarks. The experiment has two parts, (a) and (b), both shown in Figure~\ref{fig:exp2-heat}. Every model is OlmoEarth Base pretrained on its own pool of tiles and scored at step 488: one epoch for the Monotemporal pool, about 8.7 epochs for monthly+burst, and 95\% and 90\% of one epoch for mono+monthly and mono+monthly+burst.

\paragraph{(a) Sensor ladder, monotemporal tiles only.} All models in (a) are trained on the 250{,}000 monotemporal Elliot-Pretrain tiles (one acquisition per tile, $T=1$) for 488 steps. Starting from Sentinel-2 alone, we add Landsat-8/9, Copernicus DEM, ESA WorldCover and Sentinel-1 cumulatively. Sentinel-2, Landsat-8/9 and Sentinel-1 are encoder inputs (enc). DEM and WorldCover are decoder-only (dec), as in OlmoEarth: the encoder never sees them and the decoder only predicts them from the visible Sentinel-2 and Landsat-8/9 patches, so they act as auxiliary targets that are not needed at test time. We also test the opposite choice, giving DEM and WorldCover to the encoder as extra input (enc).

\paragraph{(b) Temporal partitions.} The reference in (b) is the S2+L8+DEM(dec)+WC(dec) model of (a), trained on monotemporal tiles only. We keep the model and the sensors and change only the tile pool: \emph{monthly+burst} uses only the 12{,}500 monthly and 16{,}666 burst tiles, with no monotemporal tile (29{,}166 tiles, repeated until 488 steps); \emph{mono+monthly} adds the monthly tiles to the monotemporal ones (262{,}500 tiles, 512 steps per epoch); \emph{mono+monthly+burst} uses all three (279{,}166 tiles, 545 steps per epoch). All models in (b) are scored at step 488, so they receive the same number of updates as the reference. Temporal samples are padded to 12 timesteps and share the token budget, so a sample trades spatial extent for sequence length; in the two mixed pools a monotemporal tile has one real timestep and eleven padded ones.

\paragraph{Results (a).} Figure~\ref{fig:exp2-heat} and Appendix~\ref{sec:app-exp2} give the results. Adding Landsat-8/9 alone already brings much of the gain: it improves 7 of the 9 benchmarks, ties on 2 and loses on none relative to Sentinel-2 only. Adding DEM and WorldCover as decoder-only targets on top of Landsat-8/9 improves 8 of 9 benchmarks, most on PASTIS ($+15\%$), and loses only on So2Sat ($-2\%$). Giving them to the encoder instead is worse on 5 of 9, most on So2Sat ($-14\%$). This supports treating DEM and WorldCover as decoder only targets for this benchmark suite, and shows that how a modality is added matters. The ladder is not monotone: the S2+L8+DEM model is worse than both of its neighbours on So2Sat, PASTIS, MADOS and Open-Canopy.

Sentinel-1, together with L8, DEM and WorldCover, helps on 7 of 9 benchmarks, most on TreeSatAI ($+11\%$) and BioMassters ($+7\%$), ties on 1 and loses on So2Sat ($-5\%$). It gives more consistent results than S2+L8+DEM(enc)+WC(enc), although So2Sat shows that it can also hurt, possibly on benchmarks scored with optical input only.

The multiclass segmentation benchmarks PASTIS (19 classes) and MADOS (15 classes) are the most sensitive. They gain from Landsat-8/9 ($+9\%$ and $+5\%$) and from decoder-only DEM and WorldCover ($+15\%$ and $+7\%$), lose when those layers go to the encoder ($-12\%$ and $-11\%$), and are the most affected by the temporal partitions. The scene-level classification benchmarks (EuroSAT, BrickKiln, BigEarthNet, TreeSatAI) and the two regression benchmarks (Open-Canopy, BioMassters) change by 2 to 4\% on average. 

\paragraph{Results (b).} Relative to the monotemporal reference, every temporal partition lowers the score on most benchmarks (panel b). The largest losses are on PASTIS ($-23\%$ for monthly+burst, $-45\%$ for mono+monthly, $-41\%$ for mono+monthly+burst) and MADOS ($-2\%$, $-23\%$, $-18\%$). BigEarthNet ($-2\%$ to $-10\%$) and BioMassters ($-1\%$ to $-6\%$) fall less, and the other benchmarks stay within about 6\% of the reference. The only gain is $+2\%$ on EuroSAT with monthly+burst. Monthly+burst, the only pool without monotemporal tiles, loses least on most benchmarks. This does not show that time series carry less information: samples are padded to 12 timesteps, they share the token budget, and OlmoEarth encodes the absolute acquisition date of each step. The temporal part needs more tests, for example pretraining with the years of the monthly timestamps rebased to one common year (the 12 acquisitions of a tile come from different years).

\subsection{Captions: template versus LLM}
\label{sec:exp-captions}
We compare the template with the three LLM captioners of Section~\ref{sec:exp-setup}, all writing from the same fact sheet. We took 100 stratified tiles and compared the four methods blind. Claude Opus 5.5 judged each
caption, with the image and the fact sheet as reference, on five axes, each scored from 0 to 10 (accuracy, coverage, salience,
spatial description and fluency), and counted the factual errors. We report the five scores, their mean and the errors per
tile in Table~\ref{tab:captions}. The table also lists the API cost of the captions, as a reference for users who want to generate them with an LLM. The cost is the estimated price of writing captions for all 279{,}166 tiles.

\providecommand{\capcell}[1]{\begin{tabular}[c]{@{}c@{}}#1\end{tabular}}
\begin{table}[t]
\centering
\caption{Caption quality on 100 stratified MajorTOM:Elliot Pretrain tiles. All captioners start from the
same per-tile fact sheet. A blind judge (Claude Opus 5.5) saw the tile's image, the fact
sheet and the four captions in random order. It scored each caption on five axes (0--10),
listed its factual errors, and ranked the four. Values are means over tiles, with
bootstrap 95\% confidence intervals in brackets. Costs are the measured API cost of these 100 captions (OpenRouter), scaled linearly. Best value per row in bold.}
\label{tab:captions}
\small
\setlength{\tabcolsep}{3pt}
\begin{tabular}{lcccc}
\toprule
 & \capcell{Rule-based\\template\\(no LLM)}
 & \capcell{Small LLM\\(GPT-6-luna)\\fact sheet}
 & \capcell{Large LLM\\(GPT-6.1-sol)\\fact sheet}
 & \capcell{Large LLM\\(GPT-6.1-sol)\\fact sheet + image} \\
\midrule
Accuracy & 8.74 {\scriptsize[8.51, 8.95]} & 8.73 {\scriptsize[8.54, 8.92]} & \textbf{9.39} {\scriptsize[9.27, 9.51]} & 9.32 {\scriptsize[9.19, 9.45]} \\
Coverage & 8.11 {\scriptsize[7.91, 8.29]} & 6.81 {\scriptsize[6.65, 6.96]} & \textbf{8.73} {\scriptsize[8.60, 8.86]} & 8.45 {\scriptsize[8.31, 8.59]} \\
Salience & 6.66 {\scriptsize[6.51, 6.80]} & 6.92 {\scriptsize[6.77, 7.06]} & 7.84 {\scriptsize[7.71, 7.97]} & \textbf{7.96} {\scriptsize[7.84, 8.07]} \\
Spatial & 6.92 {\scriptsize[6.54, 7.30]} & 6.45 {\scriptsize[6.09, 6.78]} & 7.42 {\scriptsize[7.03, 7.81]} & \textbf{7.45} {\scriptsize[7.09, 7.82]} \\
Fluency & 5.86 {\scriptsize[5.72, 6.01]} & 7.38 {\scriptsize[7.24, 7.51]} & 7.87 {\scriptsize[7.80, 7.94]} & \textbf{8.02} {\scriptsize[7.95, 8.09]} \\
\midrule
Mean of 5 & 7.26 {\scriptsize[7.10, 7.41]} & 7.26 {\scriptsize[7.13, 7.39]} & \textbf{8.25} {\scriptsize[8.12, 8.38]} & 8.24 {\scriptsize[8.12, 8.35]} \\
\midrule
Errors per tile $\downarrow$ & 0.65 & 0.72 & \textbf{0.40} & 0.41 \\
Mean rank (1--4) $\downarrow$ & 3.32 & 3.49 & \textbf{1.55} & 1.64 \\
Ranked first & 3\% & 2\% & \textbf{49\%} & 46\% \\
\midrule
Cost per 1,000 captions & \textbf{\$0} & \$0.43 & \$6.34 & \$9.36 \\
Cost, all 279k ELLIOT tiles & \textbf{\$0} & \$119 & \$1,769 & \$2,612 \\
\bottomrule
\end{tabular}
\end{table}

Table~\ref{tab:captions} shows the results. The template, which is free, ties with the small LLM (mean score 7.26 for both), with a similar number of errors per tile (0.65 against 0.72), so the small LLM adds nothing over it. The large LLM is better by about one point (8.25) and with fewer errors (0.40 per tile). Adding the image to the large LLM does not change the scores on average (8.24 against 8.25). Since the difference between them is not great, we use the template in the library, so that captions are free and reproducible. Because the fact sheet is plain text, anyone can also generate captions from it with the LLM of their choice and pass them to the library; the table shows what to expect from a small and a large model. Three template examples with their judge scores are in Section~\ref{sec:supplement_captions}.

There are some limitations. The comparison uses 100 tiles and one judge, which is itself a language model. It is not a downstream test: we did not train a vision language model on these captions, so we do not claim that a one point difference changes what a model learns.

\section{Discussion}
\label{sec:discussion}

\subsection{Intended Use, Limitations and Biases}
\label{sec:limitations}

\paragraph{Intended use.} The collection is meant for self-supervised pretraining of multi-sensor and
multi-temporal models, for controlled ablations over sensors and temporal regimes, and, through the task and captions,
for weak supervision and language grounding. WorldCover and the OSM entities are map products with their own errors, so they are suitable as pretraining targets and weak labels but not as reference labels for benchmarking.

\paragraph{Spatial coverage and selection.} The tiles follow Sentinel-2's acquisition footprint between
80\textdegree{} S and 84\textdegree{} N, so they include coasts and some open water, and 15.6\% of the
Monotemporal tiles lie poleward of 60\textdegree{}, where Sentinel-1 VV+VH is mostly absent. The selection
is by diversity in AlphaEarth space, not by area. The selection therefore rebalances the collection
towards rare landscape types rather than reducing the typical distance.

\paragraph{Temporal and sensor limitations.} Monotemporal images are selected to be clear (by CloudScore+), so the regime under-represents cloudy conditions, which only Monthly and Burst retain. The Sentinel-2 and Landsat-8/9 dates of a Monotemporal tile are chosen independently, and the Sentinel-1 pass can lie far from the Sentinel-2 date (90th percentile 162\,days in that regime). Sentinel-2 is top-of-atmosphere L1C, with atmospheric correction left to the user. Its 20 and 60\,m bands are replicated to 10\,m and the 30\,m Landsat and elevation layers are interpolated, so their effective resolution is coarser than the grid. The Landsat stack mixes top-of-atmosphere reflectance (bands 1 to 9, from the optical OLI) and brightness temperature in kelvin (bands 10 and 11, from the thermal TIRS). All eleven bands are stored as 32-bit integers with the same ×10,000 scale, so the thermal values are about 2.9×10$^6$ and differ by orders of magnitude from the reflectance bands; a single normalisation across the bands is wrong and per-band statistics should be used. A few tiles have no usable Landsat-8/9 acquisition (26 of 12,500 Monthly tiles, mostly in Antarctica, 18 of 250,000 Monotemporal tiles and 3 of 16,666 Burst tiles); fully missing images are stored as all-zero rasters with null product fields and \texttt{l8:nodata\_pct} equal to 1, and the others have at least 90\% nodata pixels.

\paragraph{Context layers.} OpenStreetMap completeness and naming vary strongly between regions, and 82\% of
the tiles contain at least one named entity. Names and tags are given in the local language where
available (\texttt{local\_name}). The data contain no personal information beyond what OpenStreetMap contributors have published.

\subsection{Summary of findings}

Three findings stand out. First, in our short, single-run setting the amount of unique pretraining data was not the limiting factor. A curated 25\% subset of Core-S2L2A (AlphaEarth embeddings) is better than the full corpus on three benchmarks (MADOS, BigEarthNet, So2Sat), statistically level on five and worse on PASTIS ($-3.15$ points); it is better than a random 25\% subset on four benchmarks and worse on PASTIS; and 512 random tiles are worse than the full corpus on four benchmarks. At the last checkpoint the largest gap to the full corpus is 6.57 points (MADOS, 512 tiles).
Second, how a modality is added matters. Landsat-8/9 improves 7 of 9 benchmarks and loses on none. DEM and WorldCover as decoder-only targets improve 8 of 9 (up to $+15\%$ on PASTIS), whereas giving them to the encoder is worse on 5 of 9 ($-14\%$ on So2Sat, $-12\%$ on PASTIS, $-11\%$ on MADOS). Sentinel-1 helps on 7 of 9 benchmarks, particularly on benchmarks with radar inputs such as TreeSatAI and BioMassters. The Monthly and Burst partitions lower most scores, a result confounded by padding and the token budget, so the temporal part needs more tests.
Third, a rule-based template already produces captions that tie with a small LLM (mean score 7.26 for both), a large LLM adds about one point (8.25) and lowers the errors per tile from 0.65 to 0.40, and adding the image to the prompt does not change the scores; since captions are generated from plain-text fact sheets, anyone can use the LLM of their choice.
The experiments are short (0.07\% to 6.4\% of the official schedule) and use one pretraining run per configuration, so we read them as indicative and not as a ranking of data or design choices.

\subsection{Limitations of the experiments}

The runs from the experiments use 0.07\% to 6.4\% of the official OlmoEarth schedule, each configuration is pretrained once, and encoders are frozen with one model family (OlmoEarth Base), so the results describe this early-training regime and not converged models. The confidence intervals in Experiment~1 are over the ten checkpoints of one run, include no pretraining-seed variation and are not corrected for multiple comparisons, and the three probe seeds change the nearest-neighbour scores by at most 0.1 points (So2Sat) and the linear-probe scores by at most 0.43 points (MADOS). The experiments also have confounds: Core-S2L2A (L2A) and Elliot-Pretrain (L1C) differ in processing level and are used in different experiments, the curated subset depends on one embedding model (AlphaEarth), and the temporal partitions are confounded by padding, the shared token budget and the span of years of the monthly tiles; all temporal models are scored at step 488, so the monthly+burst model sees its 29{,}166 tiles about 8.7 times, whereas mono+monthly and mono+monthly+burst stop before the end of their single epoch (at 488 of 512 and of 545 steps, 95\% and 90\% of it). Finally, the captions are judged on 100 tiles by one language model, and no vision-language model was trained on them. We therefore read the results as indicative, not as a ranking of data, sensors or captioners.

\section{Conclusions}
\label{sec:conclusion}

We presented MajorTOM:Elliot Pretrain, a collection of 279{,}166 tiles selected by clustering AlphaEarth embeddings of the Major-TOM land grid. Each tile holds Sentinel-2 L1C, Landsat-8/9, Sentinel-1 RTC, Copernicus DEM and ESA WorldCover on one $1056\times1056$-pixel, 10\,m grid, together with cloud masks, ERA5 weather and OpenStreetMap context, in three disjoint temporal regimes (Monotemporal, Monthly and Burst) released as three TACO subdatasets of Cloud Optimized GeoTIFFs, whose Parquet metadata can be filtered before any pixel is read and whose layers can be downloaded and combined independently. Checks on the metadata of every tile and on the pixels of 300 random tiles show that all layers share one grid, that the metadata match the pixels, and that the cloud masks agree with the independent CloudScore+ score (Spearman 0.84 for Monthly and 0.73 for Burst). Moreover, text is not stored: the \texttt{majortom-elliot-tasks} library generates sensor fact sheets, tasks and captions from the data.

The experiments with OlmoEarth Base suggest three points. First, selection by embedding diversity can shrink a pretraining corpus with little loss: in ten-epoch runs, an AlphaEarth-curated 25\% subset of Core-S2L2A was not worse than the full corpus on 8 of 9 benchmarks and was better than a random 25\% subset on four, although it was worse on PASTIS. Second, how a layer is used matters as much as whether it is included: Landsat-8/9 improved or matched Sentinel-2-only pretraining on every benchmark. Third, DEM and WorldCover helped as decoder targets and hurt as encoder inputs. Fourth, Sentinel-1 helped mainly TreeSatAI and BioMassters when it was also given at test time. Fifth, a free rule-based template matched a small LLM, a large LLM was about one point better, and users can write captions with any LLM from the same fact sheets.

These conclusions are limited: all runs use at most 6.4\% of the official OlmoEarth schedule with one pretraining run per configuration, and the captions were scored by a single LLM judge, not by training a vision-language model on them. Future work should (i) repeat the curation comparison with longer schedules and several pretraining seeds, (ii) train on the temporal regimes without padding to a fixed length, so that Monthly and Burst can be compared fairly, (iii) train a vision-language model on the generated tasks to test the language layer downstream, and (iv) add new layers, such as Sentinel-2 L2A or higher resolution imagery, as further subfolders and tables, together with a loading throughput benchmark that quantifies the access efficiency of the format.

\section{Data Access}
\label{sec:data_access}

The MajorTOM:Elliot Pretrain collection is publicly available on Source Cooperative at \url{https://source.coop/major-tom/elliot-pretrain}. It is distributed under the Creative Commons Attribution-ShareAlike 4.0 license (CC-BY-SA 4.0), except \texttt{osm.parquet}, \texttt{admin.parquet} and the \texttt{facts/} folder, which hold OpenStreetMap names and geometries and remain under ODbL-1.0 (\textcopyright{} OpenStreetMap contributors), as declared in the \texttt{COLLECTION.json} of each sub-dataset and in the dataset card. The Sentinel-1 data from the Planetary Computer (CC-BY-4.0) and the ERA5 weather values are compatible with these licences. Each sub-dataset is a self-contained TACO v3 folder, and its metadata revision is recorded in \texttt{ml:metadata\_revision} in \texttt{COLLECTION.json}.

The collection is plain Parquet tables and COGs (Table~\ref{tab:layout}), so it can be read with pip-installable, general-purpose libraries: the metadata with PyArrow, pandas, Polars or DuckDB and the pixels with GDAL or rasterio.
Two runnable notebooks in the \texttt{majortom-elliot-tasks} repository cover metadata queries, loading every layer and a streaming PyTorch DataLoader with parallel fetching. The TACO reader (\texttt{taco.ml}), from the fork at \url{https://github.com/OscarPellicer/taco} of \url{https://github.com/asterisk-labs/taco}, adds a single object per sub-dataset, metadata filtering before any pixel is read, on-demand decoding into arrays and checks of the collection contract.

\FloatBarrier
\bibliographystyle{unsrtnat}
\bibliography{references} 

\clearpage

\appendix
\label{sec:supplement}

\section{Design Rationale}
\label{sec:supplement_design}

This subsection adds the mechanics behind the three design choices named in
Section~\ref{sec:design-rationale}, beyond what is already stated in
Sections~\ref{sec:design-rationale}--\ref{sec:layout}.

\paragraph{Tile size.} The tile side of $1056 = 2^5\cdot3\cdot11$ pixels is divisible by the common ViT patch sizes (8, 16 and 32,
giving 132, 66 and 33 patches per side) and halves cleanly five times (528, 264, 132, 66, 33), so patch grids
and multi-scale features need no cropping or padding. It also splits into $3\times3$ internal COG tiles of
352 pixels ($11\times32$), so a reader can fetch one ninth of a tile and every block stays aligned for GPU
decoding. Two further constraints shaped the layout: fork-safe streaming into a PyTorch
\texttt{DataLoader}, and GPU-native decompression via nvCOMP.

\paragraph{Uniform 10\,m pixel grid.} Without a shared grid, a training batch would
mix incompatible shapes ($1056^2$ for Sentinel-2, $352^2$ for Landsat at its native
resolution) and \texttt{torch.stack} would fail, forcing per-batch resampling that
dominates wall time. Resampling once at creation means batches collate with no
interpolation cost at load time. Sensors finer than 10\,m remain supported as long as
their spacing divides evenly into 10\,m (5\,m, 2.5\,m, 1.25\,m): switching resolution
then reduces to nearest-neighbour sampling with an integer pixel stride, a
memory-view operation rather than an interpolation.

\paragraph{Separating the index from the pixels.} Each row in the Parquet index
carries an \texttt{internal:gdal\_vsi} column encoding a GDAL virtual file-system
path that points directly at the corresponding COG, rather than embedding the pixel
blob in the index itself. Any Parquet-capable engine (PyArrow, Polars, DuckDB, Arrow
bindings in R or Julia) can therefore evaluate filters on column statistics and skip
row groups that do not match \emph{before} a single byte of imagery is touched; only
the \texttt{gdal\_vsi} paths of surviving rows are passed to GDAL, which opens the
COGs through its VSI layer using HTTP range requests and reads only the bytes needed.
Sample selection therefore costs time proportional to the number of matching tiles,
not to the size of the archive. Table~\ref{tab:major-tom-vs-elliot} summarises the
resulting differences from Major-TOM.

\begin{table}[hb]
	\caption{Summary of changes between Major-TOM and MajorTOM:Elliot Pretrain.}
	\centering
	\small
	\setlength{\tabcolsep}{4pt}
	\begin{tabular}{p{2.3cm}p{4.6cm}p{4.6cm}}
		\toprule
		 & Major-TOM & MajorTOM:Elliot Pretrain \\
		\midrule
		Storage format & GeoTIFFs embedded in Parquet columns & Cloud-Optimised GeoTIFF via TACO \\
		Row groups & One row per row group & Up to 10{,}000 rows per row group \\
		Band handling & One column per band, mixed resolutions & All bands at 10\,m in a single COG \\
		Pixel dimensions & Variable, not aligned to 32 & 1056 $\times$ 1056 \\
		Georeferencing & Per-band, mixed grids & Single MGRS-aligned CRS + affine \\
		Compression & LZW (CPU only) & ZSTD + PREDICTOR=2 (3 for float layers) + INTERLEAVE=BAND \\
		Internal tiling & 1024 $\times$ 1024 & 352 $\times$ 352 px \\
		Index design & Parquet index + parquet data files & Parquet index + separate COG files \\
		Pixel access & Full row-group deserialisation & GDAL VSI range requests \\
		Multi-sensor join & Manual, per-dataset & Row-aligned layers, shared tile id \\
		\bottomrule
	\end{tabular}
	\label{tab:major-tom-vs-elliot}
\end{table}

\section{Global Grid Construction}
\label{sec:supplement_grid}

Every tile in the global grid is assigned a Sentinel-2 MGRS identifier, a native UTM
CRS, and a pixel-aligned affine geotransform as follows. First, $\approx$5 million
points are generated at 10\,km spacing between 80\textdegree S and 84\textdegree N
(the extent of the standard MGRS/UTM tiling scheme), with points within 15\,km of the
antimeridian removed to avoid cells straddling the 180\textdegree{} boundary. Each
point is queried against the Sentinel-2 catalog (2024--2025 imagery) and classified
by how many MGRS tiles cover it, from one to six; UTM zone overlaps make this number
grow at high latitude. For points covered by more than one tile, a
10.56\,km $\times$ 10.56\,km window centred on the point is projected
into each candidate tile's native CRS independently (adjacent tiles may not share a
UTM zone), and the candidate whose most recent acquisition footprint covers the
largest share of that window is assigned. Two geotransforms are then computed for the
resulting $1056\times1056$-pixel, 10\,m window: one from the point centroid directly,
and one that snaps the window's upper-left corner to the nearest 60\,m pixel boundary
of the native Sentinel-2 grid, so that the 10\,m, 20\,m and 60\,m Sentinel-2 bands
share a common origin (Section~\ref{sec:sensors}).

\section{Sensor Layers and Compression}
\label{sec:supplement_sensors}

Table~\ref{tab:sensor-compression} gives the exact per-sensor compression parameters
referenced in Section~\ref{sec:sensors}. All five layers use ZSTD with
$352\times352$-px internal tiling and band interleaving; \texttt{DISCARD\_LSB=}$k$ zeroes the $k$
least-significant bits of a pixel before compression, bounding reconstruction error
at $2^{k-1}$ in stored units and is set per band from the sensor's own noise floor
(e.g.\ Sentinel-2's $k{=}3$ bounds error at $\pm0.0004$ reflectance; TIRS's heavier
$k{=}10$ still bounds error at
$\pm0.05$\,K against a 0.3--1.7\,K intrinsic noise floor). WorldCover applies neither
differencing nor a bit trim, since both would corrupt categorical class codes. The Burst Landsat-8/9 rasters carry no bit trim, and the Monthly and Burst elevation rasters are stored as Float32.

\begin{table}[h]
	\caption{Per-sensor compression parameters. All COGs use ZSTD and
	$352\times352$-px internal tiles. For the elevation layer the first value is Monotemporal and the second Monthly and Burst; the Burst Landsat-8/9 rasters carry no bit trim.}
	\centering
	\small
	\begin{tabular}{lrrrl}
		\toprule
		Sensor & Bands & Dtype & Predictor & \texttt{DISCARD\_LSB} \\
		\midrule
		Sentinel-2 L1C & 13 & UInt16 & 2 & 3 (all bands) \\
		Landsat-8/9 OLI-TIRS & 11 & UInt32 & 2 & 3$\times$9 (OLI), 10$\times$2 (TIRS) \\
		Copernicus DEM GLO-30 & 1 & Int16 / Float32 & 2 / 3 & 2 / none \\
		ESA WorldCover v200 & 1 & UInt8 & 1 & none (categorical) \\
		Sentinel-1 RTC & 2 & Float32 & 3 & mantissa truncated to 12 bits \\
		\bottomrule
	\end{tabular}
	\label{tab:sensor-compression}
\end{table}

\section{The Elliot-Pretrain Collection}
\label{sec:supplement_collection}

\paragraph{Spatial selection.} AlphaEarth Foundations (AEF) \citep{brown2025alphaearth} produces embeddings on
the unit hypersphere $S^{63}$ via a von Mises--Fisher bottleneck; because the
per-tile embedding used here is the spatial \emph{average} of per-pixel
embeddings over the $1056\times1056$ patch, this breaks the unit-norm
constraint, so every embedding is renormalised to unit norm before clustering,
and tiles whose embedding norm is near zero (open ocean, permanent nodata) are
discarded beforehand. Clustering itself is a two-level spherical $k$-means
hierarchy run on GPU (FAISS): the first level partitions the $\approx$2.5M
embeddings into $K_1{=}530$ coarse clusters, and the second refines each coarse
cluster into $K_2{=}528$ sub-clusters, giving $530\times528=279{,}840$ fine
clusters; the point closest to each fine-cluster centroid becomes its
representative.

\paragraph{Monthly: greedy circular day-of-year spacing.} For each tile, the full
Sentinel-2 archive (2015--2026) is filtered to acquisitions with \emph{zero}
nodata pixels in the $1056\times1056$ window -- no partial-fill strategy is
used. Among the survivors, the highest cloud-free-score candidate seeds the
selection, and each subsequent pick is the candidate maximising the minimum
circular day-of-year (DOY) distance to every already-selected candidate, until
12 are chosen. Circular distance wraps at the year boundary (a late-December
date counts as close to an early-January one), and the 12 picks are not
constrained to one calendar year -- only coverage of the annual cycle is
enforced, not alignment to a specific year's acquisitions.

\paragraph{Burst: best 30-day window, then greedy linear spacing.} A 30-day
window is slid across the full archive at a 5-day step; each window is scored
by the product of its candidate count and its candidates' mean cloud-free
score, and the highest-scoring window is kept. This score is not date-aware, so
the chosen window can fall in any year and typically lands in the tile's local
dry season rather than astronomical summer. Within the selected window, 6
timesteps are then chosen by the same greedy-spacing logic as Monthly, but with
\emph{linear} rather than circular DOY distance, since a 30-day window never
wraps the year boundary.

\paragraph{Landsat-8/9 matching cascade.} For each Sentinel-2 date already
selected (Monthly or Burst), a Landsat-8/9 counterpart is searched through four
progressively relaxed levels; the first level returning at least one candidate
is used (Table~\ref{tab:l8-cascade}).

\begin{table}[h]
	\caption{Landsat-8/9 matching cascade. The first level with a candidate wins.}
	\centering
	\small
	\begin{tabular}{lll}
		\toprule
		Level & Temporal tolerance & Score requirement \\
		\midrule
		N1 & $\pm8$ days & $>0.8$ \\
		N2 & $\pm16$ days & $>0.6$ \\
		N3 & $\pm30$ days & $>0.4$ \\
		N4 & any date & none \\
		\bottomrule
	\end{tabular}
	\label{tab:l8-cascade}
\end{table}

\section{Cloud Masks, Reanalysis Context, Sentinel-1 and OpenStreetMap}
\label{sec:supplement_auxiliary}

\paragraph{Sentinel-1 product.} The Sentinel-1 layer described in
Section~\ref{sec:sensors} is the Planetary Computer radiometrically terrain corrected product,
cut on the tile grid, with a residual offset to Sentinel-2 of about 1 to 1.5 pixels (Section~\ref{sec:validation}) that is
not corrected: no co-registration shift is applied (\texttt{s1:shift\_applied\_px} is always $[0,0]$)
and no resampling is needed unless the UTM zone of the pass differs from that of the tile (\texttt{s1:resampled}).
Pixels driven negative by the cubic warp in that case are clamped to zero and counted in
\texttt{s1:clamped\_negative\_px}.

\paragraph{OpenStreetMap selection for the fact sheet.} The hydrated \texttt{osm.parquet} keeps small features so that
tasks such as small-object detection can use them. The fact sheet mentions only the main named features, chosen by these
rules.
\begin{itemize}
\item Polygons below 4{,}000\,m$^2$ and lines below 200\,m are dropped.
\item Points are kept only for a short list of roles (settlements, peaks, bays, lighthouses, aerodromes, railway stations,
ferry terminals), plus schools, hospitals, clinics, places of worship and markets in tiles with fewer than ten other features.
\item Coarse locators that fill the tile (95\% or more of its area), such as a region, sea or country, are dropped. So are
generic buildings below 10{,}000\,m$^2$ and construction sites, which date a description to a month.
\item Features with the same name and subtype are merged into one object, and the 18 covering the largest share of the
tile are kept.
\end{itemize}

\section{Text generation}
\label{sec:language}

The collection does not store captions. It stores the data needed to write them, and a code library
(\texttt{majortom-elliot-tasks}, MIT licence \cite{majortom_elliot_tasks}) builds fact sheets, questions, answers and
captions from the tile layers when they are needed. The only text-side data stored are the precomputed per-tile facts in
the top-level \texttt{facts/} folder (Table~\ref{tab:layout}), which let tasks be built without reading pixels. The
wording is chosen by rules and a seed, so it changes at every epoch. The library uses only the public collection.

\paragraph{Fact sheet.} For each tile, the library collects everything measured about it into a fact sheet. From the
sample metadata it takes the place and administrative names; from the Sentinel-2 and Landsat-8/9 images it computes
indicators of vegetation (NDVI), water, built-up surfaces, snow and burn scars, and the Landsat-8/9 brightness
temperature; from DEM and WorldCover it takes relief and land-cover shares. From the context layers it takes the named
OpenStreetMap features and administrative units, the cloud masks (so that cloudy pixels are ignored), the ERA5 weather on
the acquisition date and the Sentinel-1 backscatter. Every fact is tagged with the sensor that supports it, and a fact
whose sensor is not given to the model is removed, so the model is never asked about something it cannot see, such as a
temperature without Landsat-8/9. Positions are given in words on a $3\times3$ grid (north-west, centre, and so on)
instead of coordinates. For Monthly and Burst tiles the sheet also holds facts that no single image shows: the yearly
cycle of vegetation (peak and low month, number of green-up cycles), changes in water and snow extent, abrupt change
between two dates (for example a burn), which images are too cloudy to use, and the weather over the period.

\paragraph{Tasks.} From the fact sheet and the OpenStreetMap geometry the library creates 16 families of tasks
(Table~\ref{tab:tasks}). A task is only created if the layers it needs are available to the model. The wording of each
task is drawn at random but reproducibly from the tile, the family and the epoch, out of a bank of phrasings per family;
80\% of the phrasings are used for training and 20\% are held out for evaluation, so an evaluation phrasing is never
seen in training. Positions are integers on a $0$--$1000$ grid, and a derived answer (\texttt{vegetated\_fraction},
\texttt{disturbance}) states its evidence, an array of numbers, before the answer. A PyTorch \texttt{Dataset} returns, for
each tile, the image arrays of the requested sensors and a short conversation of a few tasks; the same setting decides
which arrays are loaded and which tasks are asked.

\begin{table}[tbp]
\centering
\caption{The 16 task families of \texttt{majortom-elliot-tasks}. ``Requires'' lists the layers that must be available to
the model; \emph{series} means a stack of more than one acquisition (Monthly, Burst); -- means the family is built from
the metadata and OpenStreetMap alone.}
\label{tab:tasks}
\small
\begin{tabular}{lll}
\toprule
Family & Answer shape & Requires \\
\midrule
\texttt{caption} & prose & -- \\
\texttt{dominant\_cover} & word & WorldCover \\
\texttt{which\_present} & choice & -- \\
\texttt{locate\_point}, \texttt{locate\_bbox} & point, box & -- \\
\texttt{features\_all} & several points & -- \\
\texttt{ndvi\_grid} & grid & S2 \\
\texttt{cover\_grid} & grid & WorldCover \\
\texttt{vegetated\_fraction} & number, from an evidence array & S2 \\
\texttt{acquisition} & record (date, solar elevation) & -- \\
\texttt{relief} & number & DEM \\
\texttt{peak\_greenness}, \texttt{green\_up\_count}, \texttt{clearest\_frame} & choice & S2, series \\
\texttt{ndvi\_series} & number array & S2, series \\
\texttt{disturbance} & choice, from an evidence array & S2, series \\
\bottomrule
\end{tabular}
\end{table}

\paragraph{Scoring.} Each answer is scored from 0 to 1 by a rule that depends on its shape, and an answer that cannot be
parsed in the declared shape scores 0.
\begin{itemize}
\item Words and choices: exact match after normalising case and punctuation.
\item Numbers and number arrays: correct within a tolerance per quantity (relief 10\,m or 10\%, NDVI 0.05, vegetated
fraction 10 percentage points, solar elevation 2\textdegree).
\item Points: a hit inside the reference object's box, or within 100 grid units of the reference point; boxes:
intersection over union; several points: F1 after a one-to-one match within 100 grid units.
\item Grids: share of correct cells. Records: key by key. Free text: token F1 against the reference.
\item Derived answers: the evidence and the answer are scored separately, and the reported score is the answer's.
\end{itemize}

\paragraph{Caption template.} The \texttt{caption} family is written by a rule-based template (version 2) that reads only
the fact sheet. It is a program with fixed rules that fills ready-made sentences with the measured facts and uses no
language model. The caption has one paragraph per sensor group (location and layout, optical, thermal, radar, terrain,
and cloud and weather), and a paragraph is dropped when its sensor is not given to the model. Numbers are turned into
named fractions (28\% becomes ``a little under three-tenths'', and a zero share is written as ``none''); only the
sheet's own terms are used (Landsat-8/9 gives a brightness temperature, not a surface temperature, and relief under
40\,m is not described); up to six named OpenStreetMap features are chosen across kinds (transport, water, settlements,
green areas, landmarks) and placed in compass words. Every statement has several paraphrases, sentences are reordered
if the order does not matter, and a seed picks the wording, so the same tile and seed always give the same text and
other seeds give other wordings of the same facts. A checker re-reads every proportion in a caption against the fact
sheet and reports any that the facts do not support. Users can instead pass their own captions, for example written by
a language model from the same fact sheet. Section~\ref{sec:exp-captions} compares the two options, and
Appendix~\ref{sec:supplement_captions} shows template examples.

\section{Template Caption Examples}
\label{sec:supplement_captions}

Three captions written by the template for the tiles of the caption comparison (Section~\ref{sec:exp-captions}), chosen to
span a regime, a landscape and a level of OpenStreetMap coverage. The scores are those of the blind judging, in the order
accuracy, coverage, salience, spatial description, fluency (0--10), followed by the number of errors and the rank of the
template among the four captions compared in that judging (template, GPT-6.1-sol text, GPT-6.1-sol image, GPT-6-luna text).
The judge's remark on each caption is given below it. The template varies its wording with the seed, so other seeds give
other phrasings of the same facts.

\paragraph{Edinburgh (monotemporal, urban, OpenStreetMap-rich; scores 9/9/7/9/6, 1 error, rank 3 of 4).}
\begin{quote}\small\itshape
This tile lies in the area of City of Edinburgh, Scotland, United Kingdom. Its land belongs to the island of Great
Britain. Slightly less than three-tenths of the scene is built-up land; tree cover a little under three-tenths, grassland
a little over a quarter and cropland roughly a tenth. The landscape is heavily modified by human activity. A bay, Firth
of Forth, is in the north and north-east; Fisherrow Sands, a beach, lies in the north-east. Dalkeith Country Park lies in
the east and south-east; Holyrood Park lies in the north-west and north. The Grange, a suburb, lies in the north-west and
west. City of Edinburgh Bypass, a trunk road, runs east-west through the south-west, east and south.

A little under two-fifths of the scene is vegetated, with dense vegetation over a small share of the area. Roughly a
quarter of the tile is open water; a small part of the area is unvegetated and bright in the shortwave infrared (bare
ground or sealed surfaces). Greenness peaks in the east and is lowest in the north-east, while water is concentrated in
the north-east.

On a separate Landsat-8/9 overpass the at-sensor brightness temperature averages about 1\,\textdegree C, from 0 to
3\,\textdegree C across the tile, warmest in the north-east and coolest in the west.

Sentinel-1 C-band backscatter shows canopy-like volume scattering over around a fifth of the tile, bright built-up-like
double-bounce returns over roughly a fifth of the tile, and smooth specular surfaces over a small share of the area.
Backscatter is brightest in the north-west and darkest in the north-east.

The ground lies between 0 and 292\,m, a relief of 292\,m, with most of the ground near the low end.
\end{quote}
Judge: ``with most of the ground near the low end'' is not supported by the sheet beyond a 72\,m median, a mild
overstatement. Overall, the two best captions are accurate, fluent and well located; this one is correct but stiff and adds
an unsupported terrain claim.

\paragraph{Yakutia (monthly, taiga under snow; scores 9/8/6/6/7, 0 errors, rank 4 of 4).}
\begin{quote}\small\itshape
This tile lies in the area of Oymyakonsky Ulus, Sakha Republic, Russia (two local administrative units, named in Cyrillic
in the original, are omitted here). Slightly more than seven-tenths of the area is tree cover, moss and lichen about an
eighth and grassland about a tenth. It is near-pristine, with very little human modification.

No part of the tile reaches the vegetated threshold, and the vegetation index is weakly positive on average. Open water
covers a sliver of the tile, and snow or ice covers almost all of the scene.

On a separate Landsat-8/9 overpass the at-sensor brightness temperature averages about $-32$\,\textdegree C, from $-36$ to
$-29$\,\textdegree C across the tile. The south is warmest and the north-east coolest.

C-band radar from Sentinel-1 shows canopy-like volume scattering over a little under a tenth of the scene, bright
built-up-like double-bounce returns over a small share of the tile and smooth specular surfaces over a sliver of the
scene. Backscatter is brightest in the east and darkest in the north-west.

Elevation ranges from 874 to 1{,}454\,m, a relief of 579\,m.
\end{quote}
Judge: ``built-up-like'' could mislead in a pristine snowy tile, though it faithfully paraphrases the sheet. Overall, all
four captions are largely accurate and use the compass little; this one is stiff and gives the radar fractions
trivia-like weight.

\paragraph{Botswana (burst, shrubland, OpenStreetMap-sparse; scores 9/6/6/4/6, 0 errors, rank 4 of 4).}
\begin{quote}\small\itshape
This tile lies in the area of Kweneng District, Botswana. Almost all of the area is shrubland. The landscape is
near-pristine, with very little human modification. Mapped features, mostly unnamed, include many tracks, a path and a
stream.

No part of the tile reaches the vegetated threshold, and the vegetation index is weakly positive on average. There is no
open water; all of the area is unvegetated and bright in the shortwave infrared (bare ground or sealed surfaces); a
burn-scar signature appears over just over a sixth of the tile.

Landsat-8/9 thermal data from another date give a brightness temperature of about 18\,\textdegree C on average, ranging
from 16 to 20\,\textdegree C. The north-east is warmest and the south-west coolest.

Sentinel-1 C-band backscatter shows canopy-like volume scattering over slightly more than a third of the area and smooth
specular surfaces over a very small part of the area; no bright built-up-like double-bounce returns are detected.
\end{quote}
Judge: the burn-scar share (18\%) is acceptable as ``just over a sixth'', and the radar share (36\%) as ``slightly more
than a third''. In a burst tile the elevation paragraph is absent by design, because the template drops it when the
relief is under 40\,m (26\,m here). Overall, another caption alone locates the stream and the track and describes the
human features, while the others give no image-based layout.

\section{Metadata Schema}
\label{sec:supplement_schema}

Table~\ref{tab:schema} lists the columns of the metadata tables of the release, as declared in the
\texttt{COLLECTION.json} of each sub-dataset. Every image-level table is joined to its tile through \texttt{internal:parent\_id}
(\texttt{children.parquet} links each layer folder to its tile).
The Burst tile table additionally carries \texttt{window:start\_date}, \texttt{window:end\_date},
\texttt{window:span\_days} (14--29 days between the first and last image) and \texttt{window:n\_candidates}, which describe the selected
30-day window, and \texttt{window:origin}, which records whether the tile is an original tile or a geographic-neighbour replacement.

{\footnotesize
\setlength{\tabcolsep}{4pt}
\begin{longtable}{
  >{\raggedright\arraybackslash}p{2.6cm}
  >{\raggedright\arraybackslash}p{4.7cm}
  >{\raggedright\arraybackslash}p{7.0cm}}
\caption{Metadata columns of the release.\label{tab:schema}}\\
\toprule
Table & Columns & Content \\
\midrule
\endfirsthead
\toprule
Table & Columns & Content \\
\midrule
\endhead
\bottomrule
\endfoot
tile (\texttt{sample}) & \texttt{majortom:code\_10km}, \texttt{code\_100km}, \texttt{code\_1000km} & hierarchical grid codes of the tile \\
 & \texttt{terrain:elevation} & mean elevation (m), Copernicus GLO-30 \\
 & \texttt{climate:precipitation}, \texttt{climate:temperature} & mean annual precipitation (GPM, m/year), mean annual land surface temperature (MODIS, K) \\
 & \texttt{soil:clay}, \texttt{sand}, \texttt{carbon}, \texttt{bulk\_density}, \texttt{ph} & surface soil properties as encoded by OpenLandMap: carbon in units of 5\,g/kg, bulk density in units of 10\,kg/m$^3$, pH $\times10$; null where no soil data exist (sea, Antarctica, Greenland) \\
 & \texttt{socio:gdp}, \texttt{population}, \texttt{human\_modification}, \texttt{cisi} & GDP (PPP, 2022; not per capita, aggregation undocumented), people per cell (Meta HRSL), Global Human Modification index, Critical Infrastructure Spatial Index \\
 & \texttt{admin:country}, \texttt{state}, \texttt{district} & administrative names \\
 & \texttt{ml:geometry} & tile centroid (WKB, EPSG:4326) \\
Sentinel-2 image & \texttt{s2:date}, \texttt{doy}, \texttt{id\_gee}, \texttt{product\_id}, \texttt{spacecraft}, \texttt{processing\_baseline}, \texttt{orbit\_number} & acquisition and product identifiers \\
 & \texttt{s2:score}, \texttt{nodata\_pct} & CloudScore+ clear score (0 cloudy, 1 clean), fraction of all-zero pixels \\
 & \texttt{s2:mean\_solar\_azimuth}, \texttt{solar\_zenith}, \texttt{view\_azimuth}, \texttt{view\_zenith}, \texttt{reflectance\_conversion} & illumination and viewing geometry, Sun--Earth distance factor \\
 & \texttt{ml:crs}, \texttt{geotransform}, \texttt{tensor\_shape}, \texttt{time\_start}, \texttt{time\_end}, \texttt{geometry} & raster georeferencing and footprint \\
Landsat-8/9 image & \texttt{l8:date}, \texttt{doy}, \texttt{id\_gee}, \texttt{product\_id}, \texttt{spacecraft}, \texttt{collection\_number}, \texttt{collection\_category}, \texttt{processing\_software}, \texttt{wrs\_path}, \texttt{wrs\_row} & acquisition and product identifiers \\
 & \texttt{l8:score}, \texttt{nodata\_pct}, \texttt{cloud\_cover} & clear score, fraction of all-zero pixels (1 if no coverage), scene cloud cover \\
 & \texttt{l8:sun\_azimuth}, \texttt{sun\_elevation}, \texttt{earth\_sun\_distance}, \texttt{image\_quality\_oli}, \texttt{roll\_angle} & geometry and quality \\
Elevation, land cover & \texttt{ml:tensor\_shape} & shape of the raster \\
tile (\texttt{sample}, cont.) & \texttt{id}, \texttt{internal:current\_id}, \texttt{internal:relative\_path} & tile identifier and folder path \\
 & \texttt{ml:time\_start}, \texttt{s2:product\_id}, \texttt{spacecraft}, \texttt{orbit\_number}, \texttt{score}, \texttt{nodata\_pct}, \texttt{sun:*}, \texttt{view:*} & representative Sentinel-2 scene (one of 12 or 6 in Monthly and Burst) \\
 & \texttt{osm:n\_named}, \texttt{n\_unnamed}, \texttt{n\_nodes}, \texttt{n\_admin\_units} & OSM feature counts \\
 & \texttt{worldcover:hist\_json}, \texttt{terrain:dem\_stats\_json} & WorldCover class pixel counts (in Monotemporal, class 80 means permanent water or no data), elevation statistics \\
 & \texttt{cloud:s2\_mean\_fraction}, \texttt{s2\_min\_fraction}, \texttt{l8\_mean\_fraction}, \texttt{l8\_min\_fraction} & mean and minimum cloud fraction over the acquisitions \\
 & \texttt{s1:valid\_fraction}, \texttt{valid\_fraction\_min}, \texttt{frames\_with\_data} & Sentinel-1 validity of the sample \\
Cloud mask image (S2, L8) & \texttt{ml:cloud\_fraction} & one minus the clear fraction (thick cloud, thin cloud and shadow together) \\
 & \texttt{ml:clear}, \texttt{thick\_cloud}, \texttt{thin\_cloud}, \texttt{shadow} & fraction of pixels per class \\
Sentinel-2 and Landsat-8/9 image (ERA5) & \texttt{ml:era5\_time}, \texttt{era5\_time\_is\_nominal} & ERA5 hour read, and whether the acquisition time was reconstructed \\
 & \texttt{ml:era5\_temperature\_2m\_c}, \texttt{dewpoint\_2m\_c}, \texttt{relative\_humidity\_pct}, \texttt{precipitation\_mm}, \texttt{cloud\_cover\_pct}, \texttt{wind\_speed\_10m\_kmh}, \texttt{wind\_from\_direction\_deg} & ERA5 conditions at the acquisition \\
Sentinel-1 image & \texttt{s1:s2\_frame}, \texttt{s2\_date}, \texttt{selection}, \texttt{gap\_days} & Sentinel-2 image it pairs with, selection rule, time gap in days \\
 & \texttt{s1:datetime}, \texttt{end\_datetime}, \texttt{product\_ids}, \texttt{platform}, \texttt{orbit\_state}, \texttt{relative\_orbit} & pass identifiers (null if no scene) \\
 & \texttt{s1:cover}, \texttt{valid\_fraction}, \texttt{resampled}, \texttt{shift\_applied\_px}, \texttt{clamped\_negative\_px} & footprint cover, valid pixels, resampling and clamping \\
\texttt{osm.parquet} & \texttt{osm\_type}, \texttt{osm\_id}, \texttt{category}, \texttt{subtype}, \texttt{name}, \texttt{local\_name}, \texttt{wikidata}, \texttt{wikipedia}, \texttt{tags\_json} & identity and tags of each entity \\
 & \texttt{is\_area}, \texttt{named}, \texttt{area\_m2}, \texttt{length\_m}, \texttt{area\_in\_tile\_m2}, \texttt{length\_in\_tile\_m}, \texttt{clipped}, \texttt{center\_lon}, \texttt{center\_lat}, \texttt{geometry} & size, clipping and location (geometry in the tile's UTM CRS) \\
\texttt{admin.parquet} & \texttt{osm\_type}, \texttt{osm\_id}, \texttt{admin\_level}, \texttt{name}, \texttt{local\_name}, \texttt{wikidata}, \texttt{wikipedia}, \texttt{area\_in\_tile\_m2}, \texttt{clipped}, \texttt{geom\_type}, \texttt{tags\_json}, \texttt{geometry} & administrative units intersecting the tile \\
\end{longtable}
}

\section{Validation}
\label{sec:validation}

We validated the release with checks on the metadata of every tile and with pixel checks on 300 randomly drawn
Monotemporal tiles (seed 0); Table~\ref{tab:validation} summarises them.

\paragraph{Tiles and Sentinel-1 pairing.} Every tile code is unique within its regime. For each of the 499{,}996 Sentinel-2 images the Sentinel-1 frame
paired with it records the date of that image, in 100\% of cases.

\paragraph{Grid and metadata consistency.} For all 300 sampled tiles the Sentinel-2, Landsat-8/9, elevation,
land-cover and Sentinel-1 rasters share the same CRS, geotransform and shape, and the geotransform stored in the
metadata equals the file header. The nodata fractions stored in the metadata equal those measured on the pixels
for Sentinel-2 and Sentinel-1 (maximum absolute difference 0), and every land-cover value is a valid WorldCover code (checked on Monotemporal tiles, where no data is stored as 80; in Monthly and Burst, 0 marks pixels without WorldCover data).

\paragraph{Radiometric plausibility.} Across the sampled tiles the 5th to 95th percentile of the per-tile median
Sentinel-2 red reflectance is 0.04 to 0.83, and 0.012\% of valid pixel values exceed a reflectance of 1.5. The
median Landsat-8/9 brightness temperature (band 10) of a tile lies between 246 and 313\,K, and the median Sentinel-1
backscatter between $-19.0$ and $-6.8$\,dB (VV) and between $-25.2$ and $-12.9$\,dB (VH). These ranges are typical of land, snow, ice and water surfaces, whereas values far outside them, such as a reflectance above 1.5 or a temperature of 20\,K, would indicate wrong scaling. The check covers scale and units, not absolute calibration.

\paragraph{Cross-sensor co-registration.} We estimated the shift between the Sentinel-2 and Landsat-8/9 red bands
on a central $512\times512$ window by phase correlation. The median shift is 0 pixels along both axes, 81\% of the
tiles are within one pixel and 89\% within three pixels, the native Landsat resolution. We did not re-measure the
Sentinel-2 to Sentinel-1 registration; the release documents a residual offset of about 1 to 1.5 pixels, measured by phase
correlation on 3{,}362 tiles (ascending median $(+0.86, +1.12)$\,px, descending $(+1.50, +0.06)$\,px), which is not corrected.

\paragraph{Cloud masks.} The cloud masks come from OmniCloudMask, an estimator independent of the
CloudScore+ score that guided image selection. Across the 150{,}000 Monthly images the Spearman
correlation between the cloud fraction of the mask and one minus the CloudScore+ score is 0.84 (0.73 for the
99{,}996 Burst images). Of the images CloudScore+ rates as cloudy (score $\le0.2$), 98\% (Monthly) and
96\% (Burst) have a mask cloud fraction above 0.5, and of those it rates as clean ($\ge0.8$), 84\% and 85\% have a
mask cloud fraction below 0.1.

\begin{table}[h]
\centering
\caption{Validation summary. Metadata checks cover every tile, pixel checks 300 random Monotemporal tiles.}
\label{tab:validation}
\small
\begin{tabular}{lll}
\toprule
Check & Scope & Result \\
\midrule
Sentinel-1 pairing date equals Sentinel-2 date & 499{,}996 images & 100\% \\
Same CRS, geotransform and shape for all five rasters & 300 tiles & 100\% \\
Stored geotransform equals file header & 300 tiles & 100\% \\
Metadata nodata equals measured (Sentinel-2, Sentinel-1) & 300 tiles & max.\ diff.\ 0 \\
Valid WorldCover codes (Monotemporal; 0 = no data in Monthly, Burst) & 300 tiles & 100\% \\
Sentinel-2/Landsat-8/9 shift within 1 / 3 px & 300 tiles & 81\% / 89\% \\
Cloud mask vs.\ CloudScore+, Spearman & Monthly / Burst & 0.84 / 0.73 \\
\bottomrule
\end{tabular}
\end{table}

\section{Definitions of the reported dataset statistics}
\label{sec:app-defs}

All numbers in Section~\ref{sec:stats} are computed from the TACO metadata tables of the release over all rows, without reading pixels, except when a sample is stated.
\emph{Images and tiles} are row counts of the per-image and per-tile tables. \emph{Years} are the year of \texttt{s2:date}. \emph{Cloudiness} is one minus the CloudScore+ clear score
(\texttt{s2:score}); the \emph{cloud fraction} of the masks is one minus the clear fraction (thick cloud, thin cloud and shadow) of an image (\texttt{ml:cloud\_fraction}).
A Sentinel-1 image is \emph{valid} if at least 90\% of its pixels are non-zero (\texttt{s1:valid\_fraction}$\geq0.9$); a sample has \emph{no Sentinel-1} if that fraction is 0 for every image.
The \emph{gap} is the number of days between the Sentinel-1 acquisition and the Sentinel-2 date, over valid images only. A Landsat-8/9 image is \emph{absent} if \texttt{l8:nodata\_pct}$\geq0.9$.
\emph{Latitude bands} use the centroid of each tile (\texttt{ml:geometry}). \emph{Land-cover shares} sum the WorldCover pixel counts of all tiles and divide by the total.
\emph{OpenStreetMap counts} sum \texttt{osm:n\_named}, \texttt{n\_unnamed}, \texttt{n\_nodes} and \texttt{n\_admin\_units} over tiles.
The pixel-level checks of Section~\ref{sec:validation} and the per-file sizes of Section~\ref{sec:layout} use 300 Monotemporal tiles drawn at random with seed 0.
Two storage conventions matter for reuse. Landsat-8/9 bands 1--9 are top-of-atmosphere reflectance and bands 10--11 brightness temperature in kelvin, and all eleven are stored as 32-bit integers with a common scale of $10^{-4}$,
so the thermal bands are about $2.9\times10^{6}$ and each band needs its own normalization. In the Monotemporal regime, land-cover value 80 means permanent water or no data (the two cannot be separated), whereas Monthly and Burst store no data as 0.

\section{Experimental configuration}
\label{sec:app-config}

Table~\ref{tab:exp-setup} lists what differs between the two pretraining experiments and Table~\ref{tab:app-hparams} the settings they share. Both pretrain OlmoEarth Base from scratch with the official v1.2 recipe \citep{olmoearth2026v12}; everything not listed is the official value.

\begin{table}[h]
\centering\small
\setlength{\tabcolsep}{5pt}
\caption{Configuration of the two pretraining experiments.}
\label{tab:exp-setup}
\begin{tabular}{
  >{\raggedright\arraybackslash}p{3.0cm}
  >{\raggedright\arraybackslash}p{5.6cm}
  >{\raggedright\arraybackslash}p{5.6cm}}
\toprule
 & \textbf{Experiment 1} (data volume, curation) & \textbf{Experiment 2} (modalities, partitions) \\
\midrule
Pretraining data & Core-S2L2A: all 2{,}184{,}259 tiles; random 25\% (546{,}064 tiles, seed 0, no replacement);
  curated 25\% (the first 546{,}064 tiles of the AlphaEarth ordering, see below); 512 random tiles (seed 0)
  & Elliot-Pretrain: 250{,}000 Monotemporal tiles (modality ladder); monthly+burst (29{,}166 tiles), mono+monthly (262{,}500), mono+monthly+burst (279{,}166) \\
Encoder input & Sentinel-2 L2A, 12 bands & Sentinel-2 L1C, 12 bands, plus Landsat-8/9 and Sentinel-1 as stated per run (Table~\ref{tab:app-runs}) \\
Normalization & z-score, Core-S2L2A band statistics & z-score, Elliot-Pretrain band statistics (Sentinel-1: dB statistics) \\
Steps & 42{,}660 (10 epochs of 4{,}266), in two parts of 21{,}330; the 25\% subsets are repeated within an epoch
  & 488 (Monotemporal and monthly+burst), 512 (mono+monthly), 545 (mono+monthly+burst) \\
Learning-rate schedule & Cosine, warmup 1.2\% of the run; part two resumes from step 21{,}330 with the schedule recomputed over 42{,}660 steps (the rate jumps back to the cosine midpoint) & Cosine, warmup 1.2\% of the run \\
Hardware, microbatch & one A100-80GB, 64, 16 dataloader workers (2 for the 512-tile run) & four GPUs (A30-24GB, or RTX PRO 6000 Blackwell for the Sentinel-1 runs), 32 per GPU, 8 workers per GPU \\
Checkpoints evaluated & every 4{,}266 steps (10 per run) & step 488 of each run (Table~\ref{tab:exp2-scores}) \\
Probe seeds & 3 (3622, 8956, 30991), averaged & 1 (3622) \\
\bottomrule
\end{tabular}
\end{table}

\paragraph{AlphaEarth ordering of Core-S2L2A.} The curated subset takes the first 546{,}064 tiles (25\% of the 2{,}184{,}259 filtered tiles) of a fixed ordering of Core-S2L2A built from AlphaEarth embeddings. Each tile of the pool is represented by its 64-dimensional AlphaEarth embedding (2{,}163{,}209 of the 2{,}184{,}259 tiles have one; the others go to the tail below).
A five-level hierarchical $k$-means (21{,}632, 4{,}326, 865, 173 and 34 centroids) with balanced sampling across clusters (size power 0, seed 0) selects nested subsets of 1, 5, 10, 25 and 50\% of the 2{,}245{,}886 rows of the full Major-TOM table, that is 22{,}459, 112{,}294, 224{,}589, 561{,}472 and 1{,}122{,}943 tiles.
The ordering lists the tiles of the 1\% subset, then those added by the 5\% subset, and so on, with the tiles inside each added block in random order and the remaining tiles in a shuffled tail.
The curated 25\% subset is therefore the whole 10\% subset plus a random 95.4\% of the tiles added between 10 and 25\%; it is a diversity-balanced selection, not a ranking of single tiles.
The scripts are \texttt{alpha\_order.sh}, \texttt{build\_smart\_subset} and \texttt{make\_ladder\_order} of the Core-S2L2A ladder code.

\paragraph{Normalization.} Every modality is normalized per band with statistics computed on the pretraining corpus itself, because the default OlmoEarth statistics come from another corpus;
for Sentinel-1 they differ strongly. The Core-S2L2A statistics are the per-band means and standard deviations stored in the OlmoEarth normalization configuration (\texttt{computed.json}, patched for this corpus); the Elliot-Pretrain statistics are \texttt{elliot\_band\_stats.json} next to the three partitions,
computed over valid pixels of 2{,}000 Monotemporal tiles with one 352-pixel window each, and the same statistics are applied to the Monthly and Burst tiles. The transform is the z-score $(x-\mu)/\sigma$.
Sentinel-1 backscatter is first converted from linear power to dB and its statistics are VV $-11.42\pm4.79$\,dB and VH $-17.89\pm4.85$\,dB, computed over the valid pixels of 2{,}000 random Monotemporal tiles (seed 0), one 352-pixel window each, of the released layer.

\begin{table}[h]
\centering\small
\setlength{\tabcolsep}{5pt}
\caption{Pretraining hyperparameters shared by both experiments (official v1.2 Base recipe).}
\label{tab:app-hparams}
\begin{tabular}{
  >{\raggedright\arraybackslash}p{3.4cm}
  >{\raggedright\arraybackslash}p{9.6cm}}
\toprule
\multicolumn{2}{l}{\emph{Model}} \\
Architecture & Latent masked image modelling, OlmoEarth Base encoder (88.8M parameters with Sentinel-2 only; 92.0M with Landsat-8, 95.1M with DEM or Sentinel-1 added, 98.3M with DEM and WorldCover, 101.4M with Sentinel-1, DEM and WorldCover) with a shallow predictor (decoder depth 4) \\
Patch size & 1--8 (sampled per batch), stochastic depth 0.1 \\
Position encoding & Mixed 3D RoPE (base 10{,}000, temporal coordinate scale $1/30$, time in calendar days since 2000) \\
Band dropout & Random, maximum rate 0.2, on Sentinel-2 and Landsat \\
Tokenization & One band group for Sentinel-2 (12 bands, B10 dropped) and for Landsat (11 bands) \\
\midrule
\multicolumn{2}{l}{\emph{Optimization}} \\
Optimizer & AdamW (fused), lr $10^{-4}$, weight decay 0.02, gradient clipping at norm 1.0 \\
Schedule & Cosine with linear warmup over 1.2\% of the run ($8000/667{,}200$) \\
EMA & Decay $(1.0, 1.0)$, i.e.\ no target-network update \\
Precision & FSDP, bf16 parameters, fp32 gradient reduction \\
\midrule
\multicolumn{2}{l}{\emph{Objective and masking}} \\
Loss & Patch discrimination with masked negatives ($\tau = 0.1$), plus InfoNCE with weight 0.05 \\
Masking & Random-time masking with decode ratio 0.5, encode ratio 0.5, random ratio 0.5; two masked views per sample \\
\midrule
\multicolumn{2}{l}{\emph{Batching}} \\
Global batch / token budget & 512 / 2250, with sampled height and width in $\{1,\dots,12\}$ patches \\
Seed & 3622 \\
\bottomrule
\end{tabular}
\end{table}

\paragraph{Compute.} Each five-epoch part of Experiment~1 ran on one A100-80GB for 22 to 48\,h, depending on the load of the shared node. The Experiment~2 runs with in-training evaluation of PASTIS took
2.5\,h on four A30 GPUs (mono+monthly+burst); the Sentinel-1 runs, without in-training evaluation, took 11 to 13\,min on four RTX PRO 6000 Blackwell GPUs.
Evaluating one checkpoint takes about 1\,h on one A30.

\section{Handling of modalities}
\label{sec:app-modalities}

\paragraph{Pretraining.} Table~\ref{tab:app-pretrain-mod} gives, for each input, how it enters the model and what happens to missing data.
In the temporal partitions every sample is padded to a frame of 12 time steps; absent steps are filled with a missing value and masked, and a batch draws one spatial extent, so under the token budget a sample trades
spatial extent for time steps (with Sentinel-2, Landsat-8/9, DEM and WorldCover as inputs, 12 time steps fit up to $9\times9$ patches and only 6 at $12\times12$ patches; with patches of up to 8 pixels the largest crop is $96\times96$ pixels).

\begin{table}[h]
\centering\small
\setlength{\tabcolsep}{4pt}
\caption{How each pretraining input is prepared. Statistics are those of the pretraining corpus (Appendix~\ref{sec:app-config}).}
\label{tab:app-pretrain-mod}
\begin{tabular}{
  >{\raggedright\arraybackslash}p{2.1cm}
  >{\raggedright\arraybackslash}p{1.9cm}
  >{\raggedright\arraybackslash}p{7.6cm}
  >{\raggedright\arraybackslash}p{3.2cm}}
\toprule
Input & Role & Preparation & Missing data \\
\midrule
Sentinel-2 & encoder & Elliot-Pretrain: L1C top-of-atmosphere, Core-S2L2A: L2A; both go through the same Sentinel-2 slot, with B10 dropped (12 bands) and no conversion between levels; per-band z-score & none \\
Landsat-8/9 & encoder & 11 bands (9 reflectance, 2 brightness temperature), per-band z-score & an image with $\geq90\%$ no data is treated as absent \\
Sentinel-1 & encoder & VV and VH, linear power to dB ($10\log_{10}$, floor $10^{-6}$), per-band z-score with the dB statistics & no-data pixels are set to the band mean (zero after z-scoring); the 6.6\% of tiles without a pass keep an image that is entirely mean-filled, no image is borrowed from another tile \\
DEM & decoder target (encoder input in the \emph{enc} run) & Copernicus GLO-30 in the elevation slot, per-band z-score & none \\
WorldCover & decoder target (encoder input in the \emph{enc} run) & ESA WorldCover v200 class codes, normalized as in OlmoEarth (categorical) & none \\
Time & all & timestamps $(\mathrm{day},\mathrm{month}-1,\mathrm{year})$ give a month embedding and the calendar-day coordinate of the 3D RoPE & padded steps masked \\
\bottomrule
\end{tabular}
\end{table}

\paragraph{Benchmarks.} Table~\ref{tab:app-bench} lists, for each benchmark, what is given to the model. Three rules apply to all of them.
(i)~Every optical input goes through the same 12-band Sentinel-2 slot and is standardized with the benchmark's own per-band mean and standard deviation;
L1C and L2A data are not converted into each other, and the processing level is whatever the benchmark ships. We infer it from the band list, since only L1C products carry band B10: L1C for EuroSAT and BrickKiln (13 bands) and L2A-type for BigEarthNet and TreeSatAI (12 bands); for So2Sat, PASTIS, MADOS and BioMassters the files do not settle it.
(ii)~A band that a benchmark lacks is copied from the spectrally nearest band, together with its statistics.
(iii)~Single-date benchmarks give one time step with a fixed default date (1 July 2020); PASTIS gives its 12 monthly images and BioMassters 4 monthly images.
A model with a Sentinel-1 pathway is scored with Sentinel-1 and Sentinel-2 together on PASTIS, TreeSatAI and BioMassters, and with optical input only, Sentinel-1 left out, on the other benchmarks.
A model without a Sentinel-1 pathway is scored with optical input only.
DEM and WorldCover are never given at evaluation.

\begin{table*}[h]
\centering\scriptsize
\setlength{\tabcolsep}{3pt}
\caption{Benchmarks. ``Bands imputed'' lists bands copied from a neighbouring band. Caps are the numbers of train/validation/test samples used when the split is subsampled (fixed seeded draw, identical for all models).}
\label{tab:app-bench}
\begin{tabular}{
  >{\raggedright\arraybackslash}p{2.0cm}
  >{\raggedright\arraybackslash}p{3.3cm}
  >{\raggedright\arraybackslash}p{3.6cm}
  >{\raggedright\arraybackslash}p{1.5cm}
  >{\raggedright\arraybackslash}p{2.4cm}
  >{\raggedright\arraybackslash}p{1.9cm}}
\toprule
Benchmark (classes) & Input & Bands imputed & Time steps & Probe & Metric \\
\midrule
So2Sat (17) & S2, 10 bands & B01 from B02, B09 from B8A, B10 from B11 & 1 & KNN, $k=20$ & accuracy \\
EuroSAT (10) & S2, 13 bands (L1C) & none (B10 dropped) & 1 & KNN, $k=20$ & accuracy \\
BrickKiln (2) & S2, 13 bands (L1C) & none (B10 dropped) & 1 & KNN, $k=20$ & accuracy \\
BigEarthNet (43) & S2, 12 bands (L2A) & none & 1 & KNN, $k=20$ & micro-F1 \\
TreeSatAI (15, multi-label) & S2 12 bands; S1 VV, VH & none; S1 ratio band dropped & 1 & KNN, $k=20$ & macro-F1; 10000/2000/5044 \\
PASTIS (19) & S2 (and S1) time series & as in the OlmoEarth loader & 12 & linear, lr 0.1 & mIoU \\
MADOS (15) & S2 & B6 as mean of B5 and B7, B9 as B8A, B10 as mean of B8A and B11 & 1 & linear, lr 0.01 & mIoU \\
Open-Canopy (regression) & SPOT 6/7 red, green, blue, NIR at 1.5\,m, 128-pixel crop & the other 8 bands from these four & 1 & linear, lr $10^{-3}$ & RMSE (m); 4000/1000/2000 \\
BioMassters (regression) & S2 10 bands (cloud-probability band dropped); S1 VV, VH (ascending) & B01, B09 & 4 & linear, lr $10^{-3}$ & RMSE (Mg\,ha$^{-1}$); 3000/868/1500 \\
\bottomrule
\end{tabular}
\end{table*}

\section{Evaluation protocol}
\label{sec:app-eval}

After pretraining the encoder is frozen and the patch tokens are mean-pooled; the evaluation patch size is 4. Classification benchmarks use a $k$-nearest-neighbour classifier with $k=20$.
Segmentation and regression use a linear probe trained for 50 epochs with AdamW at the learning rate and batch size of Table~\ref{tab:app-bench} (batch size 128 for MADOS, 8 for PASTIS, Open-Canopy and BioMassters),
with cross-entropy loss (segmentation) or masked mean-squared error (regression). The test metric is that of the probe after epoch 50: validation is computed only once, at the end of training, so no epoch is selected.
Across probe seeds the nearest-neighbour benchmarks differ by at most 0.01 points (So2Sat 0.1) and the linear probes by at most 0.43 points (MADOS).
Inputs are standardized per band with the benchmark's own training-split mean and standard deviation (\texttt{norm\_stats\_from\_pretrained=False}, \texttt{norm\_method=standardize});
for TreeSatAI, Open-Canopy and BioMassters these are estimated once from a fixed sample of 96 training tiles (3 frames per tile for multi-frame benchmarks, no-data excluded) and cached.
We report mIoU rather than the framework's primary metric for MADOS (micro-F1, equal to overall accuracy) and micro-F1 rather than macro-F1 for BigEarthNet.
EuroSAT, BigEarthNet, BrickKiln and MADOS are read from the OlmoEarth research-benchmark copies of GeoBench and MADOS, and So2Sat from the GeoBench copy; TreeSatAI, BioMassters and Open-Canopy are read from their TACO containers.

\section{Experiment 1: data volume and curation}
\label{sec:app-exp1}

The four pretraining sets, the two-part schedule and the checkpoints are in Table~\ref{tab:exp-setup}. The 25\% subsets and the 512-tile set have fewer than 4{,}266 batches per epoch, so their passes are packed into
a single dataloader epoch (otherwise the worker processes would restart every few steps), without changing the number of steps, the schedule or what the model sees.
The gaps quoted ``at the last checkpoint'' in Section~\ref{sec:exp-base} are differences of means over the three probe seeds at step 42{,}660, and the other differences are means over the ten checkpoints; both are in score points (0--100) for accuracy, mIoU and F1 and in the unit of the metric for RMSE.
Figure~\ref{fig:trajectory} shows the mean over the three seeds at each checkpoint. The tables below give the statistics: Table~\ref{tab:trajectory-paired} compares each subset with the full corpus over the ten checkpoints,
and Table~\ref{tab:trajectory-final} gives the final scores with their spread over probe seeds.
\begin{table*}[h]
\centering
\tiny
\setlength{\tabcolsep}{3pt}
\caption{Paired comparison of each data size with the full corpus (100\%) over the ten checkpoints of the subset-size experiment (steps 4{,}266 to 42{,}660).
Each entry is the mean difference to the 100\% run at the same checkpoint (checkpoint scores averaged over three evaluation seeds),
its 95\% confidence interval (t distribution, 9 degrees of freedom) and the two-sided p-value of a paired t-test; * $p<0.05$, ** $p<0.01$,
without correction for the 36 comparisons. The last column compares the curated and the random 25\% subsets directly (same size and steps). For the RMSE rows a negative difference is better. Checkpoints of one run are correlated and every run
was pretrained once, so the test describes the training trajectory and does not include pretraining-seed variation.}
\label{tab:trajectory-paired}
\begin{tabular}{l cccc}
\toprule
Benchmark & Random 25\% $-$ 100\% & 512 tiles $-$ 100\% & Curated 25\% $-$ 100\% & Curated $-$ random 25\% \\
\midrule
So2Sat (Acc.) & +0.49 [-0.02, +0.99] ($p$=0.056) & -1.30 [-1.77, -0.83]** ($p$=0.000) & +0.59 [+0.02, +1.17]* ($p$=0.043) & +0.11 [-0.58, +0.79] ($p$=0.729) \\
PASTIS (mIoU) & -1.17 [-1.75, -0.59]** ($p$=0.001) & +0.04 [-0.29, +0.38] ($p$=0.789) & -3.15 [-4.46, -1.83]** ($p$=0.000) & -1.98 [-3.46, -0.49]* ($p$=0.015) \\
EuroSAT (Acc.) & +0.24 [-0.16, +0.64] ($p$=0.208) & -1.38 [-1.97, -0.79]** ($p$=0.000) & +0.47 [-0.16, +1.10] ($p$=0.126) & +0.23 [-0.08, +0.54] ($p$=0.129) \\
MADOS (mIoU) & -0.77 [-1.19, -0.34]** ($p$=0.003) & -3.93 [-5.20, -2.67]** ($p$=0.000) & +2.32 [+1.78, +2.87]** ($p$=0.000) & +3.09 [+2.35, +3.83]** ($p$=0.000) \\
BrickKiln (Acc.) & -0.56 [-0.92, -0.20]** ($p$=0.006) & +0.89 [-0.04, +1.83] ($p$=0.059) & +0.06 [-0.40, +0.52] ($p$=0.773) & +0.62 [+0.28, +0.96]** ($p$=0.003) \\
BigEarthNet ($\mu$F1) & +0.32 [+0.02, +0.62]* ($p$=0.041) & -0.02 [-0.49, +0.46] ($p$=0.941) & +0.99 [+0.71, +1.28]** ($p$=0.000) & +0.68 [+0.41, +0.94]** ($p$=0.000) \\
TreeSatAI (macro-F1) & -0.26 [-0.39, -0.13]** ($p$=0.001) & -0.33 [-0.48, -0.18]** ($p$=0.001) & -0.13 [-0.30, +0.03] ($p$=0.099) & +0.13 [+0.04, +0.22]* ($p$=0.012) \\
Open-Canopy (RMSE, m) & +0.013 [-0.010, +0.036] ($p$=0.238) & +0.019 [-0.004, +0.042] ($p$=0.090) & +0.053 [-0.025, +0.130] ($p$=0.159) & +0.040 [-0.045, +0.125] ($p$=0.316) \\
BioMassters (RMSE, Mg\,ha$^{-1}$) & +0.12 [-0.02, +0.25] ($p$=0.084) & +0.06 [-0.02, +0.14] ($p$=0.115) & +0.01 [-0.43, +0.46] ($p$=0.949) & -0.10 [-0.48, +0.27] ($p$=0.548) \\
\bottomrule
\end{tabular}
\end{table*}

\begin{table*}[h]
\centering
\scriptsize
\setlength{\tabcolsep}{3pt}
\caption{Score at the last checkpoint (step 42{,}660): mean $\pm$ standard deviation over three evaluation seeds (3622, 8956, 30991) and the 95\% confidence interval of
the mean (t distribution, 2 degrees of freedom) in brackets. The encoder is fixed, so this interval covers only the refitting of the probe or
nearest-neighbour classifier, not pretraining-seed variation. Nearest-neighbour benchmarks give identical scores for all seeds.}
\label{tab:trajectory-final}
\begin{tabular}{l cccc}
\toprule
Benchmark & 100\% & Random 25\% & 512 tiles & Curated 25\% \\
\midrule
So2Sat (Acc.) & 37.73 $\pm$ 0.00 [37.73, 37.73] & 38.74 $\pm$ 0.00 [38.74, 38.74] & 37.53 $\pm$ 0.00 [37.53, 37.53] & 38.74 $\pm$ 0.00 [38.74, 38.74] \\
PASTIS (mIoU) & 31.22 $\pm$ 0.10 [30.98, 31.46] & 29.89 $\pm$ 0.10 [29.65, 30.14] & 31.97 $\pm$ 0.07 [31.79, 32.16] & 30.21 $\pm$ 0.08 [30.02, 30.40] \\
EuroSAT (Acc.) & 79.60 $\pm$ 0.00 [79.60, 79.60] & 79.60 $\pm$ 0.00 [79.60, 79.60] & 77.40 $\pm$ 0.00 [77.40, 77.40] & 79.90 $\pm$ 0.00 [79.90, 79.90] \\
MADOS (mIoU) & 51.81 $\pm$ 0.19 [51.33, 52.28] & 50.55 $\pm$ 0.20 [50.07, 51.04] & 45.23 $\pm$ 0.07 [45.05, 45.42] & 53.74 $\pm$ 0.10 [53.49, 53.98] \\
BrickKiln (Acc.) & 85.79 $\pm$ 0.00 [85.79, 85.79] & 85.69 $\pm$ 0.00 [85.69, 85.69] & 86.79 $\pm$ 0.00 [86.79, 86.79] & 86.49 $\pm$ 0.00 [86.49, 86.49] \\
BigEarthNet ($\mu$F1) & 49.22 $\pm$ 0.00 [49.21, 49.23] & 49.57 $\pm$ 0.00 [49.57, 49.57] & 49.15 $\pm$ 0.00 [49.15, 49.15] & 50.48 $\pm$ 0.00 [50.48, 50.48] \\
TreeSatAI (macro-F1) & 18.55 $\pm$ 0.00 [18.55, 18.55] & 18.44 $\pm$ 0.00 [18.44, 18.44] & 18.28 $\pm$ 0.00 [18.28, 18.28] & 18.60 $\pm$ 0.00 [18.60, 18.60] \\
Open-Canopy (RMSE, m) & 5.568 $\pm$ 0.001 [5.565, 5.571] & 5.575 $\pm$ 0.001 [5.572, 5.578] & 5.580 $\pm$ 0.003 [5.574, 5.586] & 5.599 $\pm$ 0.001 [5.597, 5.601] \\
BioMassters (RMSE, Mg\,ha$^{-1}$) & 51.80 $\pm$ 0.01 [51.78, 51.81] & 52.07 $\pm$ 0.01 [52.04, 52.10] & 52.00 $\pm$ 0.03 [51.93, 52.07] & 52.19 $\pm$ 0.01 [52.16, 52.21] \\
\bottomrule
\end{tabular}
\end{table*}

\section{Experiment 2: sensors and temporal partitions}
\label{sec:app-exp2}

\paragraph{Pretrained models.} Table~\ref{tab:app-runs} lists the eleven models. Each is pretrained once and scored at step 488 (the monthly+burst pool of 29{,}166 tiles has only 56 steps per epoch and is repeated about 8.7 times
to reach 488 steps, matching the Monotemporal runs; mono+monthly and mono+monthly+burst have 512 and 545 steps per epoch and are scored before its end). The Sentinel-1 models use all 250{,}000 tiles, the same pool as the models without Sentinel-1.

\begin{table}[h]
\centering\small
\setlength{\tabcolsep}{4pt}
\caption{Pretrained models of Experiment~2. All use Sentinel-2 as an encoder input. dec = decoder-only target; enc = encoder input.}
\label{tab:app-runs}
\begin{tabular}{
  >{\raggedright\arraybackslash}p{4.1cm}
  >{\raggedright\arraybackslash}p{2.6cm}
  >{\raggedright\arraybackslash}p{2.0cm}
  >{\raggedright\arraybackslash}p{1.6cm}
  r}
\toprule
Model & Other encoder inputs & Decoder targets & Pool & Steps \\
\midrule
S2 only & --- & --- & Mono & 488 \\
S2+L8 & L8 & --- & Mono & 488 \\
S2+L8+DEM (dec) & L8 & DEM & Mono & 488 \\
S2+L8+DEM (dec)+WC (dec) & L8 & DEM, WC & Mono & 488 \\
S2+L8+DEM (enc)+WC (enc) & L8, DEM, WC & --- & Mono & 488 \\
S2+L8+S1 & L8, S1 & --- & Mono & 488 \\
S2+L8+S1+DEM (dec) & L8, S1 & DEM & Mono & 488 \\
S2+L8+S1+DEM (dec)+WC (dec) & L8, S1 & DEM, WC & Mono & 488 \\
monthly+burst & L8 & DEM, WC & Monthly, Burst & 488 \\
mono+monthly & L8 & DEM, WC & Mono, Monthly & 512 \\
mono+monthly+burst & L8 & DEM, WC & Mono, Monthly, Burst & 545 \\
\bottomrule
\end{tabular}
\end{table}

\paragraph{Figure~\ref{fig:exp2-heat}.} Each cell is the change of a model's score relative to a reference, $100\,(s-s_{\mathrm{ref}})/s_{\mathrm{ref}}$ for higher-is-better metrics and $100\,(s_{\mathrm{ref}}-s)/s_{\mathrm{ref}}$ for RMSE,
so that positive is always better; cells are rounded to whole percent and values that round to zero are shown as 0.
The reference is the S2-only model in panel (a) and the monotemporal S2+L8+DEM+WC (dec) model in panel (b), always on the reference model's own input (its scores are in the grey row).
The Sentinel-1 models are scored with Sentinel-1 input in the cells marked with a diamond, and the reference for those cells is the S2-only model on the Sentinel-2 version of the same benchmark.

The scores are those of step 488 of each run, with evaluation seed 3622 (Table~\ref{tab:exp2-scores}). A cell counts as ``improved'' or ``worse'' in the text if its rounded value is positive or negative, and as ``equal'' if it rounds to 0.
Averages per benchmark quoted in the text are means of the absolute values of the optical-input cells over the ten non-reference rows (seven in panel (a), three in panel (b)).
\begin{table*}[h]
\centering\scriptsize
\setlength{\tabcolsep}{3pt}
\caption{Scores behind Figure~\ref{fig:exp2-heat} with optical input, at step 488 of each run. Test split, evaluation seed 3622; 0--100 scale, RMSE for Open-Canopy (m) and BioMassters (Mg\,ha$^{-1}$). dec = decoder-only target, enc = encoder input.}
\label{tab:exp2-scores}
\begin{tabular}{lrccccccccc}
\toprule
Pretraining run & Step & So2Sat & PASTIS & EuroSAT & MADOS & BrickKiln & BigEarthNet & TreeSatAI & Canopy & BioM \\
\midrule
S2 only & 488 & 30.4 & 26.0 & 69.0 & 37.6 & 81.6 & 41.1 & 17.1 & 6.22 & 55.13 \\
S2+L8 & 488 & 31.2 & 28.5 & 70.7 & 39.5 & 86.5 & 42.5 & 17.1 & 6.23 & 54.22 \\
S2+L8+DEM (dec) & 488 & 29.3 & 25.2 & 69.4 & 36.6 & 83.9 & 41.0 & 17.1 & 6.31 & 55.20 \\
S2+L8+DEM (dec)+WC (dec) & 488 & 29.9 & 29.9 & 71.6 & 40.2 & 86.9 & 43.4 & 17.7 & 6.10 & 52.81 \\
S2+L8+DEM (enc)+WC (enc) & 488 & 26.1 & 23.0 & 69.6 & 33.6 & 81.2 & 40.6 & 17.3 & 6.43 & 54.95 \\
S2+L8+S1 & 488 & 29.4 & 23.9 & 67.1 & 36.2 & 82.9 & 41.3 & 17.0 & 6.33 & 54.59 \\
S2+L8+S1+DEM (dec) & 488 & 30.0 & 23.7 & 65.8 & 35.7 & 83.9 & 41.5 & 16.9 & 6.35 & 54.89 \\
S2+L8+S1+DEM (dec)+WC (dec) & 488 & 28.9 & 26.2 & 70.1 & 38.2 & 86.0 & 41.6 & 17.4 & 6.22 & 53.52 \\
monthly+burst & 488 & 28.5 & 22.9 & 72.8 & 39.3 & 82.9 & 42.5 & 17.3 & 6.24 & 53.28 \\
mono+monthly & 488 & 29.4 & 16.5 & 68.6 & 30.8 & 84.8 & 38.9 & 17.1 & 6.37 & 56.23 \\
mono+monthly+burst & 488 & 29.6 & 17.6 & 69.8 & 32.9 & 85.1 & 39.6 & 16.6 & 6.39 & 55.21 \\
\bottomrule
\end{tabular}
\end{table*}

\begin{table*}[h]
\centering
\setlength{\tabcolsep}{3pt}
\caption{Scores of the Sentinel-1 models with Sentinel-1 input (Sentinel-1 and Sentinel-2 together), same checkpoints and protocol as Table~\ref{tab:exp2-scores}. These are the diamond cells of Figure~\ref{fig:exp2-heat}.}
\label{tab:exp2-scores-s1}
\begin{tabular}{lrccc}
\toprule
Pretraining run & Step & PASTIS & TreeSatAI & BioM \\
\midrule
S2+L8+S1 & 488 & 26.1 & 18.7 & 52.02 \\
S2+L8+S1+DEM (dec) & 488 & 25.8 & 18.4 & 52.17 \\
S2+L8+S1+DEM (dec)+WC (dec) & 488 & 27.1 & 19.0 & 51.32 \\
\bottomrule
\end{tabular}
\end{table*}

\end{document}